\documentclass{article}

\usepackage{PRIMEarxiv}

\usepackage[utf8]{inputenc} 
\usepackage[T1]{fontenc}    
\usepackage{hyperref}       
\usepackage{url}            
\usepackage{booktabs}       
\usepackage{amsfonts}       
\usepackage{nicefrac}       
\usepackage{microtype}      
\usepackage{lipsum}
\usepackage{fancyhdr}       
\usepackage{graphicx}       
\usepackage{CJKutf8}        
\usepackage{minted}
\usepackage[most]{tcolorbox}
\usepackage{enumitem}
\graphicspath{{figures/}}     
\usepackage[version=4,arrows=pgf-filled,
textfontname=sffamily,
mathfontname=mathsf]{mhchem}
\usepackage{xcolor}

\title{ChatCFD: An LLM-Driven Agent for End-to-End CFD Automation with Structured Knowledge and Reasoning} 

\usepackage{authblk}
\author[2, $\dag$]{E Fan}
\author[1, $\dag$]{Kang Hu}
\author[3]{Zhuowen Wu}
\author[1]{Jiangyang Ge}
\author[3]{Jiawei Miao}
\author[3]{Yuzhi Zhang}
\author[4,5]{He Sun}
\author[1,6]{Weizong Wang}
\author[1,7,*]{Tianhan Zhang}

\affil[1]{School of Astronautics, Beihang University}
\affil[2]{Department of Mechanics and Aerospace Engineering, Southern University of Science and Technology}
\affil[3]{DP Technology, Beijing}
\affil[4]{College of Future Technology, Peking University }
\affil[5]{National Biomedical Imaging Center, Peking University}
\affil[6]{State Key Laboratory of High-Efficiency Reusable Aerospace Transportation Technology}
\affil[7]{Key Laboratory of Spacecraft Design Optimization and Dynamic Simulation Technology, Ministry of Education}
\affil[$\dag$]{These authors contributed equally to this work and are listed alphabetically.}
\affil[*]{Corresponding author, \texttt{thzhang@buaa.edu.cn}}

\begin{document}
\maketitle
\begin{abstract}
Computational Fluid Dynamics (CFD) is essential for advancing scientific and engineering fields, but is hindered by operational complexity, high expertise requirements, and limited accessibility.
This paper introduces ChatCFD, a LLM-driven agent system for end-to-end CFD automation. 
Powered by DeepSeek-R1/V3, a multi-agent architecture, structured OpenFOAM knowledge bases, precise error locator, and iterative reflection, ChatCFD dramatically outperforms prior systems. On 315 benchmark cases it attains \textbf{82.1\% execution success} (vs. 6.2\% MetaOpenFOAM, 42.3\% Foam-Agent) and, crucially, \textbf{68.12\% physical fidelity}—the first rigorous metric capturing whether a runnable simulation is scientifically meaningful. A dedicated \textbf{Physics Interpreter} achieves \textbf{97.4\% summary fidelity}, exposing the striking LLM gap between fluent narration and enforcing dozens of tightly coupled physical constraints in executable code. Resource-efficiency analysis highlights ChatCFD's practical advantage, achieving an average of \textbf{192.1k tokens and \$0.208 per case}—half the tokens of Foam-Agent and 1.5× cheaper than MetaOpenFOAM. Ablation studies confirm structured domain knowledge and reasoning is indispensable: removing the Solver Template DB collapses accuracy to 48\%, while the Error Locator Module proves the single most critical component. Flexibility experiments show autonomous solver selection across compressible/incompressible and steady/transient regimes (95.23\% success) and turbulence-model switching (100\% success), even on unseen configurations. By faithfully reproducing complex literature cases NACA0012 airfoil and supersonic nozzle with 60--80\% end-to-end success where baselines fail entirely, ChatCFD establishes new benchmarks for AI-driven CFD automation. Physics coupling analyses reveal higher resource demands for multi-physics-coupled cases, while LLM bias toward simpler setups introduces persistent errors. ChatCFD's modular, MCP-compatible design directly enables collaborative multi-agent networks and paves the way for scalable AI-driven CFD innovation. The code for ChatCFD is available at: \url{https://github.com/ConMoo/ChatCFD}.

\end{abstract}


\section{Introduction}

Computational Fluid Dynamics (CFD) is a cornerstone technology in diverse scientific and engineering disciplines, including aerospace \cite{slotnick2014cfd, fan2025numerical}, energy systems \cite{iranzo2019cfd, amponsah2024computational, yao2025solving}, urban environment \cite{li2021review, juan2025wind, shen2024equilibrium}, combustion \cite{posch2025turbulent, zhang2022multi,wang2024deep}, astrophysics\cite{zhang2025deep}, and biomedical applications \cite{morris2016computational, doost2016heart}. It provides indispensable tools for simulating intricate fluid behaviors, thereby enabling design innovation and scientific discovery \cite{anderson1995computational, blazek2015computational}. However, CFD implementation faces substantial barriers, requiring deep domain expertise \cite{wang2024recent} and often relying on expensive commercial software. Even experts invest significant time in tasks such as solver selection, model setup, mesh generation, boundary condition definition, and post-processing \cite{kodman2024comprehensive}. These demands, combined with high computational costs, limit CFD accessibility for smaller organizations and hinder broader innovation \cite{blazek2015computational}. Thus, there is an urgent need for automated, intuitive, and affordable CFD solutions. To meet this need, we present ChatCFD, an AI-driven agent system that streamlines CFD workflows, facilitating automated scientific discovery in fluid mechanics and engineering by enabling rapid iteration and exploration of complex flow phenomena without extensive manual intervention.

Recent progress in Artificial Intelligence (AI) has transformed the automation of sophisticated scientific processes. Large Language Models (LLMs), such as GPT \cite{achiam2023gpt}, Gemini \cite{team2023gemini}, and DeepSeek \cite{guo2025deepseek}, alongside multi-agent frameworks like MetaGPT \cite{hong2023metagpt} and AutoGen \cite{wu2023autogen}, excel in natural language processing, code generation, and reasoning via chain-of-thought techniques \cite{wei2022chain}.  Frameworks like ReAct \cite{yao2023react} synergize reasoning and acting in LLMs, while Reflexion \cite{shinn2023reflexion} uses verbal reinforcement learning to improve agent performance through self-reflection. Toolformer \cite{schick2023toolformer} teaches LLMs to self-learn tool usage. Retrieval-Augmented Generation (RAG) \cite{lewis2020retrieval} further bolsters these agents by incorporating domain-specific knowledge, reducing hallucinations, and improving domain adaptation for tasks like CFD automation.

This synergy has enabled LLM-based agents to automate CFD pipelines, from interpreting user inputs to running simulations. OpenFOAM \cite{jasak2007openfoam}, a leading open-source CFD platform, supports this trend due to its flexibility and community backing. Notable efforts include MetaOpenFOAM \cite{chen2024metaopenfoam}, which automates simulations from natural language using RAG and MetaGPT, though limited to tutorial-level cases and struggling with complex geometries; OpenFOAMGPT \cite{pandey2025openfoamgpt}, evaluating RAG-augmented LLMs for zero-shot setup and boundary adjustments, but facing difficulties in turbulence model modifications; Foam-Agent \cite{foamagent_yue25}, introducing hierarchical retrieval for dependency-aware generation, yet constrained in handling unseen configurations; and NL2FOAM \cite{dong2025fine}, fine-tuning LLMs for natural language-to-CFD translation using LoRA \cite{hu2022lora}, but reliant on domain-specific datasets and less effective for diverse physical models. Despite these advances, existing systems often fall short in end-to-end automation for complex cases without human oversight, particularly due to a fundamental challenge: LLMs lack sufficient training on niche scientific tasks like OpenFOAM setup, necessitating expert-designed architectures to bridge this gap through structured reasoning, reflection, and knowledge integration.

Existing CFD automation agents are often limited by their focus on rudimentary tasks, typically restricted to OpenFOAM tutorial-level examples, raising concerns about their generalization to complex, unseen cases. This challenge primarily stems from the scarcity of domain-specific scientific coding corpora, such as OpenFOAM setups, requiring sophisticated multi-agent system designs to effectively incorporate physical knowledge into agent architectures, thereby enhancing LLMs' capabilities in specialized scientific tool invocation. Recent representative work in this area includes SciToolAgent \cite{ding2025scitoolagent}, which develops a knowledge graph-driven agent for integrating hundreds of scientific tools across biology, chemistry, and materials science, using graph-based RAG and safety-checking to automate workflows like protein engineering and chemical synthesis. While not directly focused on CFD, its approach to tool integration and domain adaptation offers valuable insights for fluid automation agents in handling coupled physics and complex simulations. 

Three critical challenges in CFD automation agents remain unaddressed. First, incorporating domain-specific knowledge into CFD agents remains challenging. Engineering-scale simulations require comprehensive initial and boundary condition specifications, necessitating domain-specific structural thinking and seamless integration of specialized domain expertise into the automation pipeline. Second, designing effective agent frameworks, especially for future multi-MCP agent collaborations, is essential to maximize complementary capabilities. Typical issues include whether we need to contain mesh generation in the agent scope, where real-world geometries exceed basic OpenFOAM tools like \texttt{blockMesh} and require specialized integration of external meshes. Third, enabling large-scale testing and evolution of fluid agents is crucial. Current reliance on simplistic natural language descriptions often leads to ambiguous case setups and hinders scalability. In contrast, CFD literature, comprising millions of papers with detailed descriptions and standard results, remains underutilized, impeding agent evolution. These shortcomings highlight a critical gap in current CFD automation frameworks.

\begin{figure}[h!]
    \centering
    \includegraphics[width=5.5in]{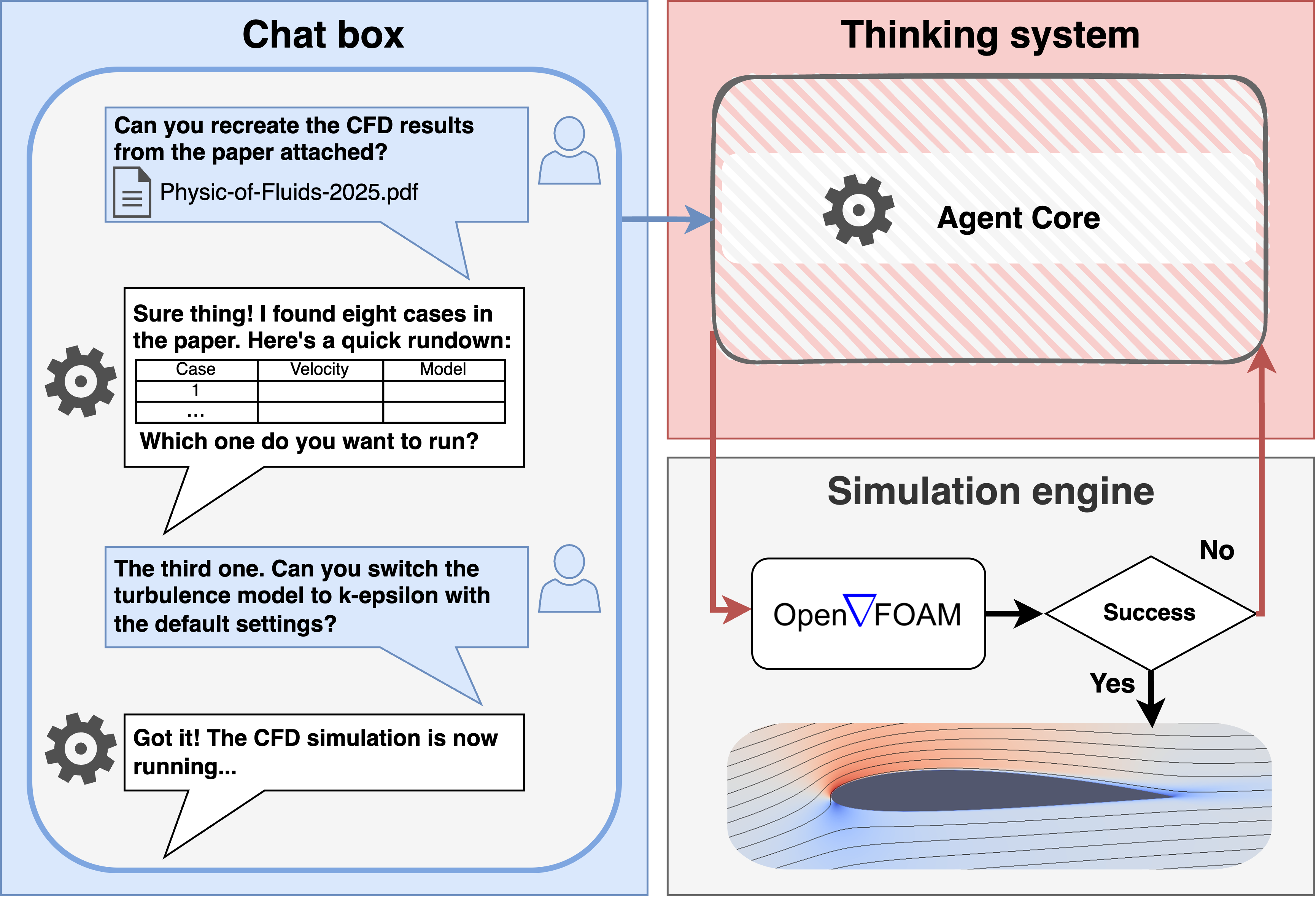}
    \vspace{8 pt}
    \caption{Overview of the ChatCFD automated agent system for streamlining CFD simulations within the OpenFOAM framework. ChatCFD enables researchers and engineers to configure and execute simulations with minimal CFD or OpenFOAM expertise. The system comprises three core components: (1) an interactive chat interface for users to input case descriptions or upload mesh files, (2) a thinking system, the core decision-making module (detailed in Figure \ref{fig_chatcfd_framework}), and (3) a simulation engine that executes cases, collects error logs, and delivers final results.}
    \label{fig_chatcfd_structure_1}
\end{figure}
To overcome these challenges, we introduce ChatCFD, an automated agent system for OpenFOAM simulations, as shown in Figure \ref{fig_chatcfd_structure_1}. ChatCFD processes multi-modal inputs (e.g., research papers, meshes) through an interactive interface, utilizing LLMs (\texttt{DeepSeek-R1}, \texttt{DeepSeek-V3}), multi-agent architecture, and OpenFOAM knowledge to manage the full workflow. It supports intricate setups via iterative refinement, handling diverse physical models and external meshes, surpassing prior agents in adaptability to unseen cases.

This paper details ChatCFD’s architecture and validates its components experimentally. The structure is as follows: Section \ref{sec:method} describes the pipeline design, Section \ref{sec:Results} presents results, and Section \ref{sec:conclusion} summarizes findings. Appendices provide examples and prompts for operational clarity.

\section{ChatCFD Pipeline Design}
\label{sec:method}

\begin{figure}[h!]
    \centering
    \includegraphics[width=5in]{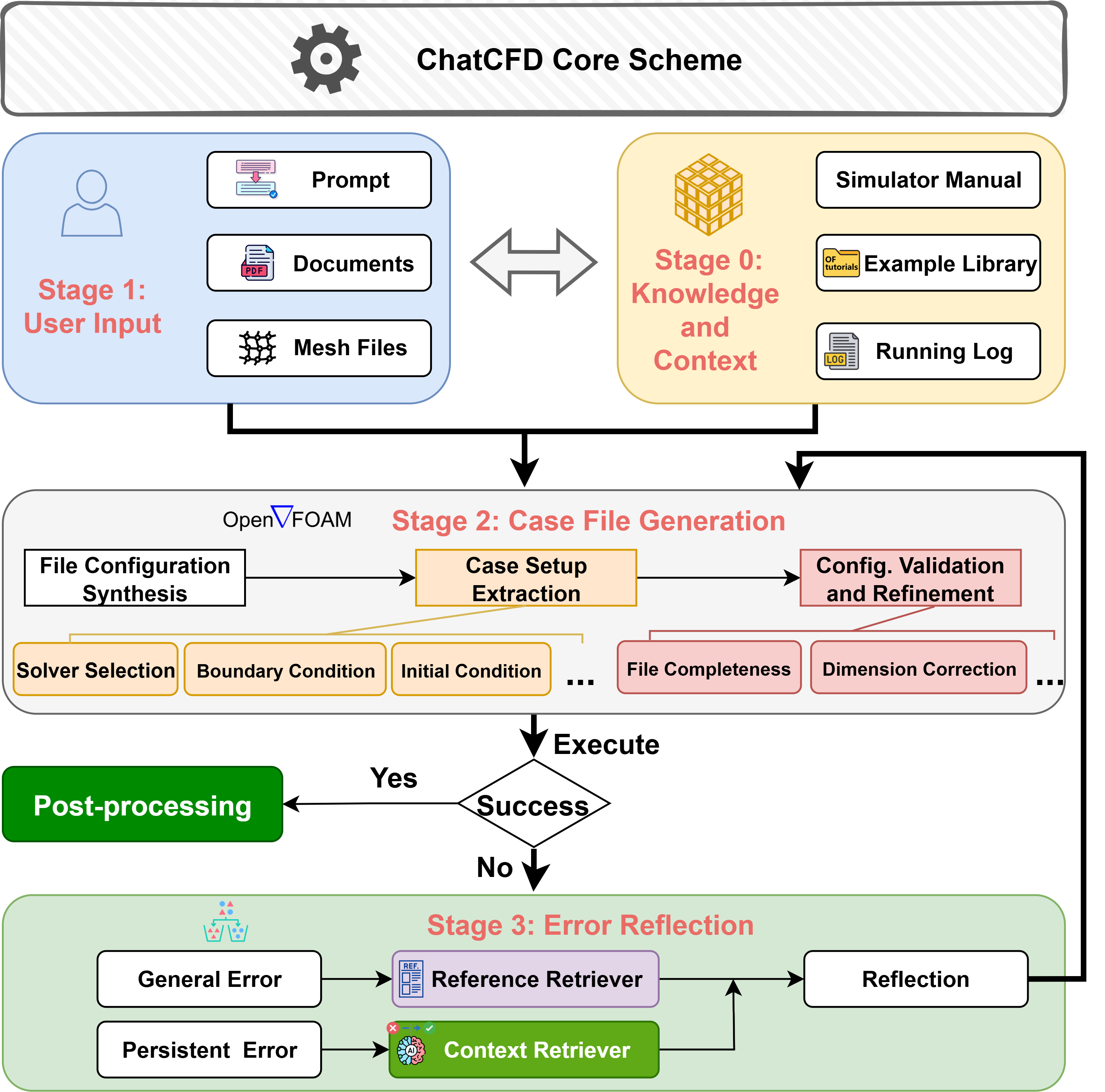}
    \vspace{8 pt}
    \caption{Architecture of the ChatCFD framework for automated CFD simulations, illustrating the four-stage workflow and agent structure. The stages are: (1) Knowledge Base Construction, creating a JSON database from OpenFOAM manuals and tutorials; (2) User Input Processing, enabling user interaction via natural language or document and mesh uploads; (3) Case File Generation, generating OpenFOAM case files using the knowledge base; and (4) Execution and Error Reflection, running simulations, converting meshes with \texttt{fluentMeshToFoam}, and resolving errors (dimension mismatches, missing files, persistent errors, general issues) using RAG-based modules \texttt{ReferenceRetriever} and \texttt{ContextRetriever}. The agent structure integrates \texttt{DeepSeek-R1} and \texttt{DeepSeek-V3} for intelligent processing, with iterative error correction.}
    \label{fig_chatcfd_framework}
\end{figure}

ChatCFD is an automated CFD agent system that leverages the OpenFOAM framework to process multi-modal user inputs, including research articles and mesh files, to configure and execute CFD simulations based on user instructions. As illustrated in Fig.~\ref{fig_chatcfd_framework}, the ChatCFD framework implements a comprehensive workflow to streamline simulation setup, execution, and analysis.

\begin{figure}[h!]
  \centering
  \includegraphics[width=0.7\linewidth]{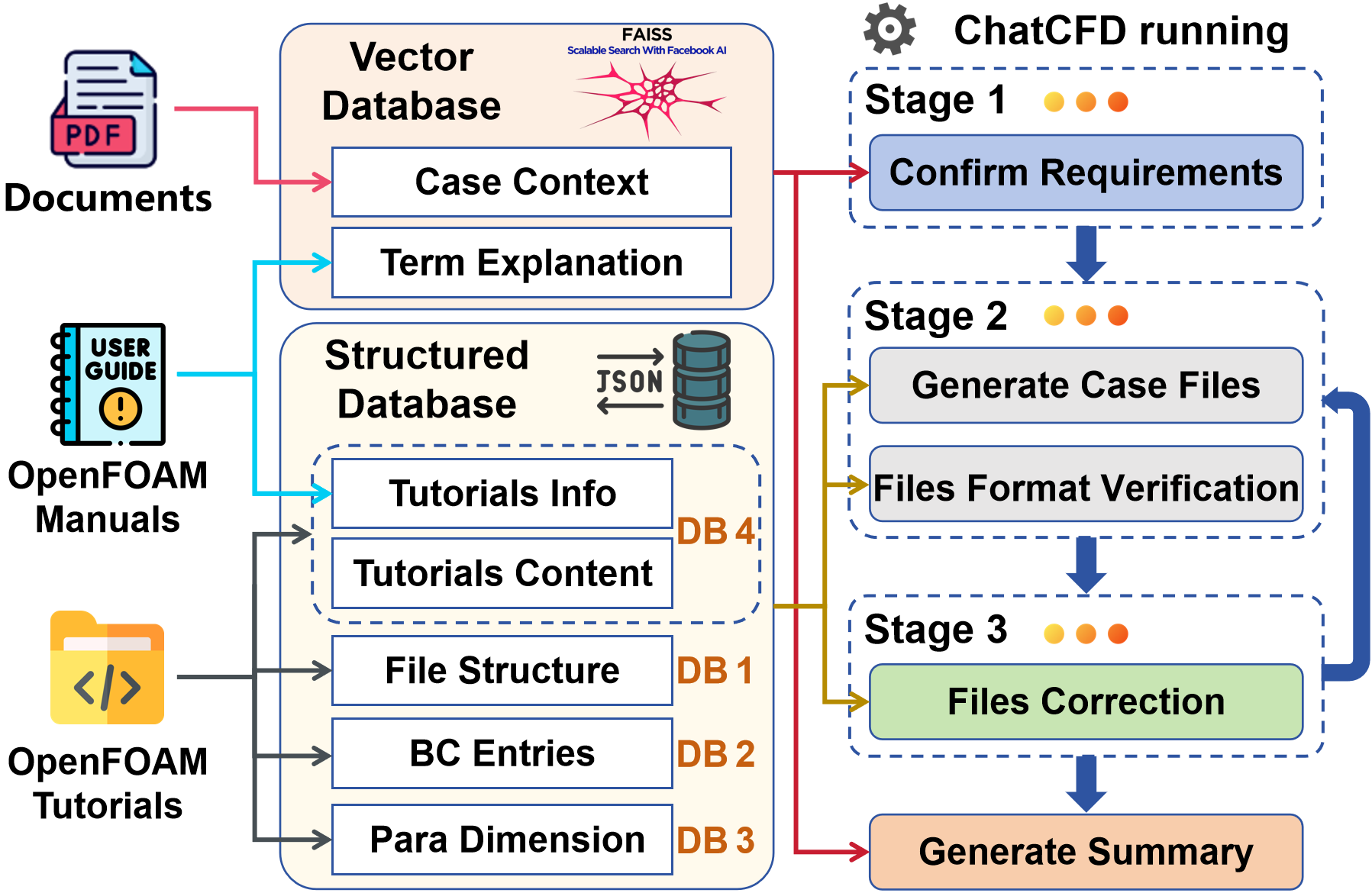}
  \vspace{0 pt}
  \caption{Based on the knowledge base constructed from the contents such as the OpenFOAM manual and OpenFOAM tutorials, and how it functions within the ChatCFD framework}
  \label{fig_db}
\end{figure}

\subsection{Stage 0: Knowledge Base Construction}
\label{sec:preprocessing_data}
The preprocessing aims to establish a foundational knowledge base for CFD tasks, performed in advance to optimize ChatCFD’s operations. Ideally, this knowledge base contains general CFD principles (e.g., numerical differentiation, discretization schemes), OpenFOAM-specific manuals, and a comprehensive example library. Fine-tuning large language models (LLMs) for general CFD knowledge is beyond the current scope; readers are referred to recent work by Dong et al.~\cite{dong2025fine}. This study focuses on preprocessing OpenFOAM manuals and publicly accessible examples. 

\textcolor{black}{Through a systematic analysis of OpenFOAM manuals and tutorials, we construct a structured JSON database that records key parameters such as solver configurations, turbulence models, and file dependencies, laying the foundation for accurate case setup and error correction. Additionally, a vector database is established to further enhance the understanding of OpenFOAM and to generate more precise OpenFOAM case descriptions. As shown in Fig. \ref{fig_db}, the database primarily comprises two components: a Structured Database in JSON format and a Vector Database for efficient retrieval. The Structured Database includes the following parts:}

\begin{itemize}[leftmargin=*]
\item \textcolor{black}{\textbf{File Dependency and Structure DB (db1)}: Designed to construct a hierarchical file structure model. This model is based on OpenFOAM tutorial cases and categorized primarily by solver type and turbulence model to store the necessary file structure for specific combinations. It also records the actual file structure of each tutorial case, which is utilized to construct and validate the space of permissible configurations.}
\item \textcolor{black}{\textbf{Boundary Condition DB (db2)}: Comprehensively catalogs all available boundary condition types within OpenFOAM. For each boundary condition, it precisely documents the required additional setting keywords and their configuration specifications, ensuring the completeness and validity of the boundary setup.}
\item \textcolor{black}{\textbf{Parameter Dimension DB(db3)}: Systematically records the standard dimensional configuration rules for all physical quantities in OpenFOAM. This database further defines the dimensional requirements and consistency constraints for relevant field variables under varying physical problems or specific solving environments.}
\item \textcolor{black}{\textbf{Solver Template DB (db4)}: Systematically stores the core file content and structured metadata of fundamental OpenFOAM tutorial cases. The metadata includes: the case's designated solver, turbulence model, and physical problem category. This information serves as the key basis for achieving high-accuracy classification and rapid screening of tutorial cases, thus providing precise reference templates for automated case initialization and configuration.}
\end{itemize}

\textcolor{black}{All configuration files are carefully parsed, converted to JSON format, and tagged with metadata (e.g., solver specifications and flow characteristics) for easy retrieval and application. OpenFOAM-specific header files are removed, and cases with external dependencies or auxiliary folders are excluded to maintain case autonomy and reduce the influence of irrelevant configuration elements on the LLM's response quality. Furthermore, for cases where non-uniform field definitions occupy a large volume of file content (potentially exceeding dialogue context limits), placeholders (e.g., \texttt{\_\_CELL\_COUNT\_\_}, \texttt{\_\_FACE\_COUNT\_\_}, \texttt{\_\_SCALAR\_DATA\_PLACEHOLDER\_\_}) are used to replace the non-uniform field data. This allows the LLM to focus on critical settings (such as boundary and initial conditions) while mitigating the issue of a reduced case count due to complex data.}

Initially, preprocessed manual and tutorial data are used to define file dependency relationships, which are accessed during the RAG process to enhance the agent’s error correction capabilities. To ensure robust operation, strategic case filtering is implemented by removing OpenFOAM-specific headers and excluding cases with external dependencies or auxiliary folders. This maintains case independence, reducing the risk of degraded LLM response quality due to irrelevant configuration elements.
For cases where non-uniform field definitions occupy significant file content and may exceed the dialogue context limits of large models, placeholder tokens such as \texttt{\_\_CELL\_COUNT\_\_},\texttt{\_\_FACE\_COUNT\_\_},\texttt{\_\_SCALAR\_DATA\_PLACEHOLDER\_\_} are used to replace non-uniform field data. This approach enables the LLM to focus on critical settings, such as boundary and initial conditions, while ignoring non-uniform field configurations. As a result, it mitigates the reduction in the number of usable OpenFOAM tutorial cases caused by complex non-uniform field data.
Filtering criteria and implementation details are provided in Appendix A.
Subsequently, the example library is preprocessed through systematic extraction and organization. All case configuration files are parsed and converted into a structured JSON format, tagged with metadata such as solver specifications, turbulence model types, and flow regime characteristics. Analysis of solver and turbulence model distributions determines specific file requirements for each configuration. For instance, the \texttt{simpleFoam} solver requires configuration files including \texttt{system/fvSolution}, \texttt{system/controlDict}, \texttt{system/fvSchemes}, \texttt{0/p}, \texttt{0/U}, \texttt{constant/transportProperties}, and \texttt{constant/turbulenceProperties}. Similarly, the k-$\omega$ SST turbulence model requires \texttt{system/fvSolution}, \texttt{system/controlDict}, \texttt{system/fvSchemes}, \texttt{0/p}, \texttt{0/U}, \texttt{0/k}, \texttt{0/omega}, \texttt{0/nut}, \texttt{constant/turbulenceProperties}, and \texttt{constant/transportProperties}.



    
    
    
    
    




\subsection{Stage 1: User Input Processing}
\label{sec:interacting}
The initial stage of ChatCFD features an interactive, multi-modal interface that enables users to define CFD simulations by either conversing with the \texttt{DeepSeek-R1} model or uploading documents and mesh files, built using the Streamlit Python framework. This interface leverages the preprocessed knowledge base to facilitate accurate case specification with minimal CFD expertise. The knowledge base, comprising OpenFOAM manuals, tutorial cases, and a structured JSON database of solver configurations and turbulence models, informs the system’s natural language processing and case extraction, ensuring robust and context-aware interactions. The workflow comprises four key phases:

\begin{itemize}[leftmargin=*]
\item \textbf{Input Submission}: Users can upload research articles or describe the case through dialogue to provide the basis for case analysis.
\item \textbf{Case Extraction}: The system extracts and catalogs CFD cases from uploaded documents or conversation inputs, identifying solver configurations and physical models, and presents them in a structured format with unique identifiers (e.g., Case 1, Case 2).
\item \textbf{Case Selection}: Through natural language interaction, the system guides users to select a target case, provides detailed specifications for verification, and confirms the selection.
\item \textbf{Mesh Integration}: The system assists users in uploading mesh files, currently supporting Fluent format \texttt{.msh} files. For OpenFOAM default meshes, users must specify the path to the corresponding polyMesh directory.
\end{itemize}

Upon completing this workflow, ChatCFD aggregates three essential components for downstream processing, as shown in Fig.~\ref{fig_chatcfd_framework}:

\begin{itemize}[leftmargin=*]
\item \textbf{Case Documentation}: The system retains the full research article or conversation details, providing access to all technical specifications, including solvers, boundary conditions, and simulation parameters.
\item \textbf{Case Specification}: The selected case is recorded with detailed metadata, including solver types, numerical schemes, physical models, and source references, ensuring precise setup.
\item \textbf{Mesh Data}: Robust file transfer and path management ensure reliable delivery of mesh data to subsequent stages.
\end{itemize}

\subsection{Stage 2: Case File Generation}
\label{sec:initiating_case_files}

The CFD engineer layer, powered by the \texttt{DeepSeek-R1} and \texttt{DeepSeek-V3} models, initializes OpenFOAM case files through a streamlined three-phase process that leverages the preprocessed knowledge base.
\subsubsection{Phase 1: File Configuration Synthesis}
The system analyzes the case description from Stage 1 (User Input Processing) to identify suitable solvers and models, thereby generating a list of required case file configurations. This involves explicitly matching dependencies for solvers, models, and related settings to facilitate accurate file configuration. The workflow proceeds as follows:
\begin{itemize}[leftmargin=*]
\item Relevant cases are retrieved based on the solver and model, yielding a corresponding file configuration list from which non-essential files (e.g., those for visualization or mesh generation) are removed. These refined cases are then input to the \texttt{DeepSeek-V3} model to produce an optimized file configuration list.
\item In the absence of direct matches, cases are aligned by solver type. For additional models, such as turbulence models, necessary files are inferred from analogous cases, synthesized into a new configuration list, verified for solver compatibility, and finalized via \texttt{DeepSeek-V3}.
\end{itemize}
This approach integrates solver and physical model structures with case synthesis, enhancing initialization precision, minimizing file omissions, and substantially improving overall case file quality.initialization.

\subsubsection{Phase 2: Case Setup Extraction}
Detailed case configurations are derived from three primary sources: the file configuration list, user prompts, and the case library. The configuration list prevents omissions or redundancies in the case structure, while user prompts enable filtering of segmented case description chunks using the \texttt{sentence-transformer/all-mpnet-base-v2} library and a similarity threshold. The \texttt{DeepSeek-R1} model implements a hierarchical three-step extraction:
\begin{itemize}[leftmargin=*]
    \item \textbf{Mesh Boundary Condition Extraction:} Boundary names and types are extracted via regular expressions and tools like pyFoam, forming a persistent dictionary for subsequent physical field generation and error handling.
    \item \textbf{Boundary Condition Validation:} Boundary conditions are identified from filtered chunks, checked for compatibility with OpenFOAM-v2406, and subjected to rigorous naming and format validation. 
    \item \textbf{Physical Field File Value Setup:} Field files (e.g., \texttt{0/p}, \texttt{0/nut}, \texttt{0/nutTilda}) are identified from the configuration list. Using prior boundary data, OpenFOAM-compliant templates are created, encompassing internal and boundary field settings differentiated by boundary names and associated keywords.
\end{itemize}

\subsubsection{Phase 3: Configuration Validation and Refinement}
For unspecified parameters, such as discretization schemes and solver algorithms in \texttt{system/fvSolution} and \texttt{system/fvSchemes}, the \texttt{DeepSeek-R1} model applies advanced reasoning grounded in CFD best practices and case-specific physics to formulate coherent configurations. This maintains parameter interdependencies and ensures alignment with the outlined flow dynamics and solver demands.

A comprehensive validation and correction mechanism further refines the generated files. The system scrutinizes structures in the \texttt{system} and \texttt{constant} directories, confirming proper dependencies and completeness through cross-references with library exemplars. Particular emphasis is placed on physical field dimensions to mitigate LLM limitations in interpreting such constraints. For example, variables like \texttt{p} and \texttt{alphat}---which differ dimensionally between compressible and incompressible flows---are calibrated according to case type; ambiguous cases retain initial values. This process curtails errors like dimensional inconsistencies, bolsters initial file robustness, and diminishes iterative corrections in Stage 3.

\subsection{Stage 3: Error Correction and Reflection}
\label{sec:running_case}
Stage 3 of the ChatCFD pipeline employs a modular architecture for automated error correction and iterative refinement in OpenFOAM case configurations. The following modules facilitate robust error handling:
\begin{itemize}[leftmargin=*]
\item \textbf{\texttt{ReferenceRetriever}}: This module retrieves reference files from the preprocessed OpenFOAM knowledge base, matching solver and model specifications from benchmark tutorial cases. If no exact match is found, it prioritizes solver-compatible files. To balance guidance and prompt complexity, two reference files are selected to aid error detection and correction, leveraging the LLM’s few-shot learning to enhance ChatCFD’s accuracy in error locator and resolution.
\item \textbf{\texttt{ContextRetriever}}: This module compiles current case configurations into a structured JSON format, providing detailed file content and modification trajectories over a specified period, tailored to other modules’ needs. It supports targeted reflection and correction, particularly for errors involving physical coupling (e.g., pressure-density interactions).
\item \textbf{Error Locator Module}: Comprising two components: (1) \texttt{DeepSeek-V3} rapidly identifies suspicious files based on error messages, reducing the analysis burden; (2) \texttt{DeepSeek-R1} performs detailed reasoning to pinpoint erroneous files by comparing their content against the case library, leveraging its advanced inference capabilities for precise error location.
\item \textbf{Error Correction Module}: This module addresses file interdependencies in OpenFOAM through a three-step process: (1) \texttt{DeepSeek-V3} identifies files requiring coordinated modifications to resolve coupling-related errors; (2) \texttt{DeepSeek-R1} proposes actionable corrections by integrating error details, related file content, simulation requirements, and benchmark tutorial cases; (3) \texttt{DeepSeek-V3} applies these corrections, ensuring compliance with OpenFOAM’s file format standards.
\item \textbf{Reflection Module}: Activated during persistent errors, this module collects error messages and file modification histories, formulating reflections in the format: ''For situation A, I considered B but overlooked C; next, I will apply D to resolve the issue.'' These reflections are stored as reflection blocks within a \texttt{<reflexion>} tag in the reflection history, enhancing iterative error correction by integrating with error locator and correction modules.
\end{itemize}

Errors detected during simulation are categorized into two types:
\begin{itemize}[leftmargin=*]
\item \textbf{General Errors}: These, comprising over 70\% of runtime errors, include incorrect keywords, formatting issues, or floating-point errors. Correction involves a streamlined workflow: suspicious file detection (\texttt{DeepSeek-V3}), error confirmation (\texttt{DeepSeek-R1}), retrieval of related tutorial files (\texttt{ReferenceRetriever}), proposal of corrections, and file modification. For missing file errors, \texttt{DeepSeek-V3} directly generates the required file using reference cases, followed by dimensional validation to ensure quality.
\item \textbf{Persistent Errors}: ChatCFD leverages short-term memory (recent error messages and file modification histories) and long-term memory (reflection histories) to address recurring issues. Reflection histories, stored as structured insights, enhance the system’s ability to adapt and resolve complex errors iteratively.
\end{itemize}

\subsection{Stage 4: Post-Processing}
\textcolor{black}{To enhance the transparency and user intelligibility of the computation process, ChatCFD integrates a dedicated \textbf{Physics Interpreter} in the fourth stage of the workflow. This component, implemented using the DeepSeek-R1 model, automatically produces a technical summary immediately after the case is successfully generated and deemed executable. The output is presented in Markdown format, supplemented by structured expressions such as tables and file tree diagrams. Its content systematically covers multiple dimensions, including fundamental information, physical problem background, special configuration settings, file structure, and key file contents, thereby comprehensively ensuring the transparency of the simulated physical process for the user.}

\begin{CJK*}{UTF8}{gbsn}

\end{CJK*}

\subsection{LLM Architecture and Configuration}
\label{sec:genernal_setups}
ChatCFD employs a dual-model architecture, integrating the \texttt{DeepSeek-R1} and \texttt{DeepSeek-V3} large language models (LLMs) to execute a comprehensive CFD workflow for case generation and error correction. Each model is strategically deployed based on its strengths to optimize performance across Stages 2 and 3.

The \texttt{DeepSeek-R1} model, with its superior text comprehension and reasoning capabilities, is utilized for complex, high-stakes tasks. In Stage 2, it extracts critical simulation parameters (e.g., boundary conditions, initial conditions) from case descriptions and generates complete initial case files. In Stage 3, it pinpoints error-causing files, proposes detailed correction strategies, and conducts reflection on persistent errors. However, \texttt{DeepSeek-R1}’s propensity for hallucinations—such as inserting unnecessary Markdown formatting, which can disrupt error correction—poses challenges. Despite mitigation efforts using prompt engineering and \texttt{pydantic} for structured outputs, these issues persist. Additionally, its inference time increases significantly with longer prompts, making it less efficient for routine tasks.

The \texttt{DeepSeek-V3} model excels in instruction following and rapid response, making it ideal for structured, low-complexity tasks. In Stage 2, it performs quick file structure validation, while in Stage 3, it identifies suspicious or missing files and generates corrected file content based on provided instructions. To address \texttt{DeepSeek-V3}’s limited reasoning capacity, tasks are simplified into structured formats (e.g., providing configuration file lists to narrow analysis scope), and the model is prompted to justify its actions (e.g., explaining why a file is suspicious). This enhances its effectiveness while maintaining computational efficiency.
This dual-model approach leverages \texttt{DeepSeek-R1}’s reasoning for complex CFD tasks and \texttt{DeepSeek-V3}’s efficiency for structured operations, ensuring robust performance across diverse case types (e.g., benchmark tutorial cases, literature-derived cases) while minimizing computational costs.

ChatCFD maintains a comprehensive log for each CFD case, capturing all question-answer interactions and actions taken during reflection-based error correction. These logs are invaluable for analyzing the behavior of the \texttt{DeepSeek-R1} and \texttt{DeepSeek-V3} models within the OpenFOAM framework, particularly in identifying limitations in handling complex CFD configurations. The recorded reflection iterations, which document corrective actions for persistent errors, provide critical insights into scenarios where ChatCFD’s performance is suboptimal, guiding targeted improvements in the pipeline. Additionally, a concise summary of reflections for recurring errors is maintained, offering a deeper understanding of ChatCFD’s behavior in complex scenarios. These logs have the potential to form a novel database, serving as a valuable resource for enhancing LLM-based agents in addressing high-complexity CFD problems within the MCP framework.

The experimental validations described in Section \ref{sec:Results} were conducted on the VolcEngine platform \cite{VolcEnginePlatform}, which provides the pricing structure for token usage and computational costs. For \texttt{DeepSeek-R1}, input tokens cost \$0.00055 per 1,000 (0.004 RMB), and output tokens cost \$0.0022 per 1,000 (0.016 RMB). For \texttt{DeepSeek-V3}, input tokens cost \$0.00021 per 1,000 (0.0015 RMB), and output tokens cost \$0.00082 per 1,000 (0.006 RMB). All token consumption and cost metrics reported in Section \ref{sec:Results} are based on this pricing, enabling precise evaluation of ChatCFD’s computational efficiency across benchmark tutorial cases, perturbed variant cases, and literature-derived cases.

\section{Results and discussion}
\label{sec:Results}
The performance of ChatCFD was rigorously evaluated through a series of validation experiments encompassing three distinct categories of CFD cases: (i) 205 benchmark tutorial cases drawn from OpenFOAM tutorials and the OpenFOAM wiki, serving as standardized references for foundational validation; (ii) 110 perturbed variants, derived by systematically altering key parameters (e.g., boundary conditions, solver settings, or physical properties) in the benchmark cases to assess robustness and sensitivity; and (iii) 2 advanced literature-derived cases, directly prompted from published research articles to test real-world applicability in complex, unseen scenarios.
ChatCFD demonstrated an operational success rate of 82.1\% across the 315 benchmark and perturbed cases, with success defined as error-free configuration and execution leading to converged simulations. For the advanced literature-derived cases, success rates ranged from 60\% to 80\%, reflecting the challenges of handling intricate, domain-specific configurations without prior exposure. These results represent a substantial advancement in LLM-driven CFD automation, particularly in reducing the expertise barrier for setup and execution within the OpenFOAM framework.

\subsection{Benchmarking Execution Success Rates Across Agents}

\begin{figure}[h!]
\centering
\includegraphics[width=6in]{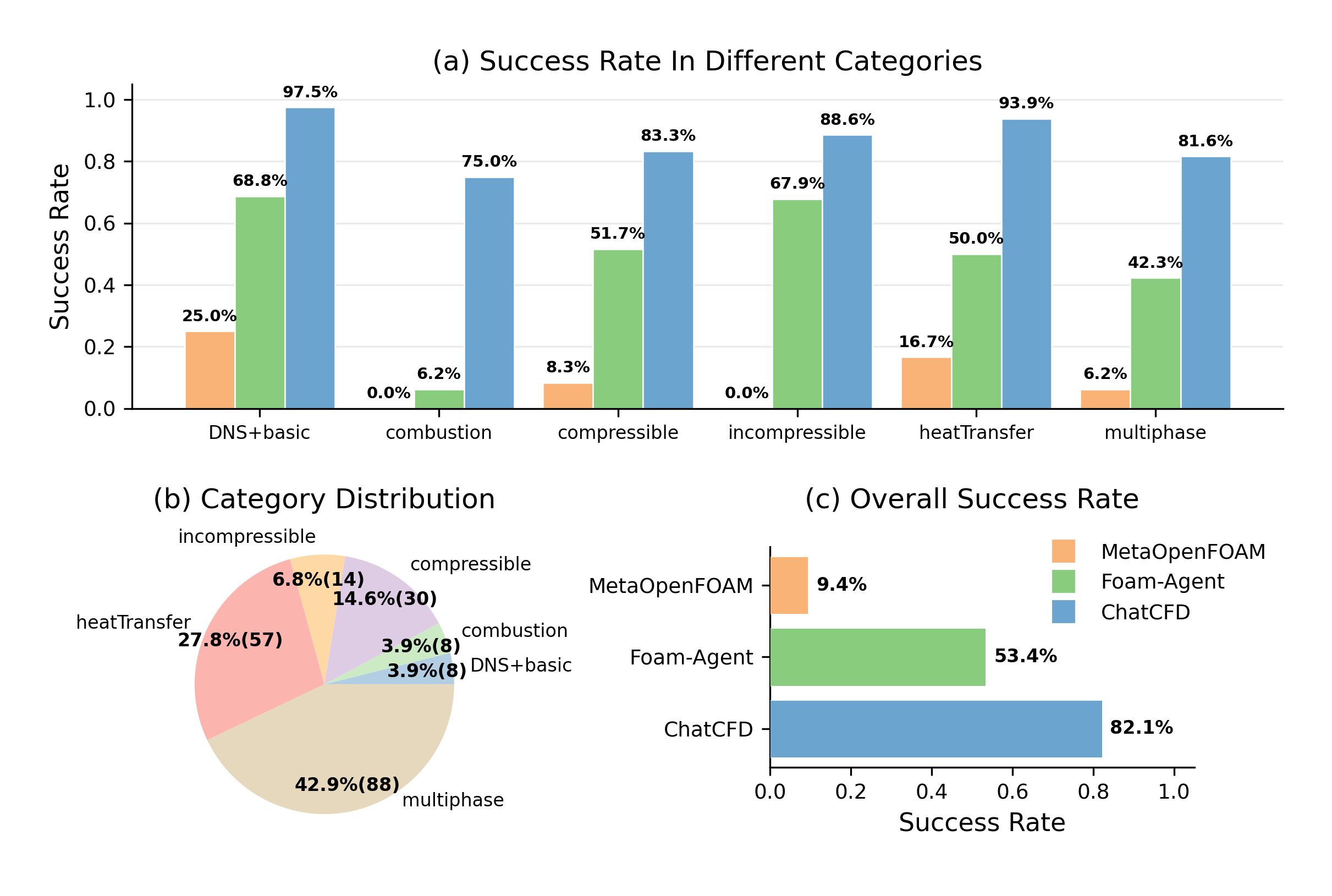}
\vspace{0 pt}
\caption{Comparison of success rates across three CFD agents (ChatCFD, MetaOpenFOAM, Foam-Agent) for 205 benchmark and 110 perturbed OpenFOAM tutorial cases. (a) Success rates by case category. (b) Distribution of test cases across categories. (c) Overall success rate comparison.}
\label{fig_205_test}
\end{figure}

To evaluate ChatCFD, a comprehensive set of test cases was curated from OpenFOAM tutorials and the OpenFOAM wiki. Cases involving complex mesh generation techniques were filtered out, resulting in 205 benchmark tutorial cases. From these, 11 cases were selected and systematically perturbed by modifying their descriptions and adjusting parameters (e.g., boundary conditions, solver settings, or physical properties), yielding an additional 110 perturbed variant cases. This approach ensures both foundational validation and robustness testing under varied conditions.
Figure \ref{fig_205_test} illustrates the performance of ChatCFD compared to MetaOpenFOAM and Foam-Agent across these cases, showing success rates by category, case distribution, and overall performance. Success is defined as the generation of error-free case files leading to converged simulations. As shown in Figure \ref{fig_205_test}(c), ChatCFD achieves an overall success rate of 82.1\%, significantly outperforming MetaOpenFOAM (6.2\%) and Foam-Agent (42.3\%). Figure \ref{fig_205_test}(a) highlights ChatCFD’s consistent advantage across case categories, demonstrating its superior performance and generalization capability.
\label{sec:OpenFOAM_tutorials_cases}

Notably, all agents exhibit lower success rates in combustion and multiphase flow categories due to their interdisciplinary complexity, which involves intricate file structures and content. For example, in the \texttt{combustionFlame2D} case, the \texttt{DeepSeek-R1} model always erroneously formulates the \texttt{thermo.compressibleGas} file by adding a \texttt{mixture} dictionary for species like \ce{O2} and \ce{H2O}, leading to persistent structural errors. This stems from an intuition that the gas is a mixture, overlooking OpenFOAM’s specific requirement that \texttt{thermo.compressibleGas} defines compressible gas thermophysical properties, while \texttt{thermophysicalProperties} is the appropriate file for \texttt{mixture} settings. To address this, ChatCFD employs two strategies: (1) retrieving and integrating relevant OpenFOAM case libraries, leveraging the LLM’s few-shot learning to adapt to specific file structures; and (2) enhancing the reflection module, which enables the system to review decision trajectories after repeated errors, identify knowledge or action discrepancies, and adopt corrected solutions. These mechanisms significantly improve ChatCFD’s handling of challenging cases, boosting success rates in complex scenarios.

\begin{figure}[h!]
\centering
\includegraphics[width=6in]{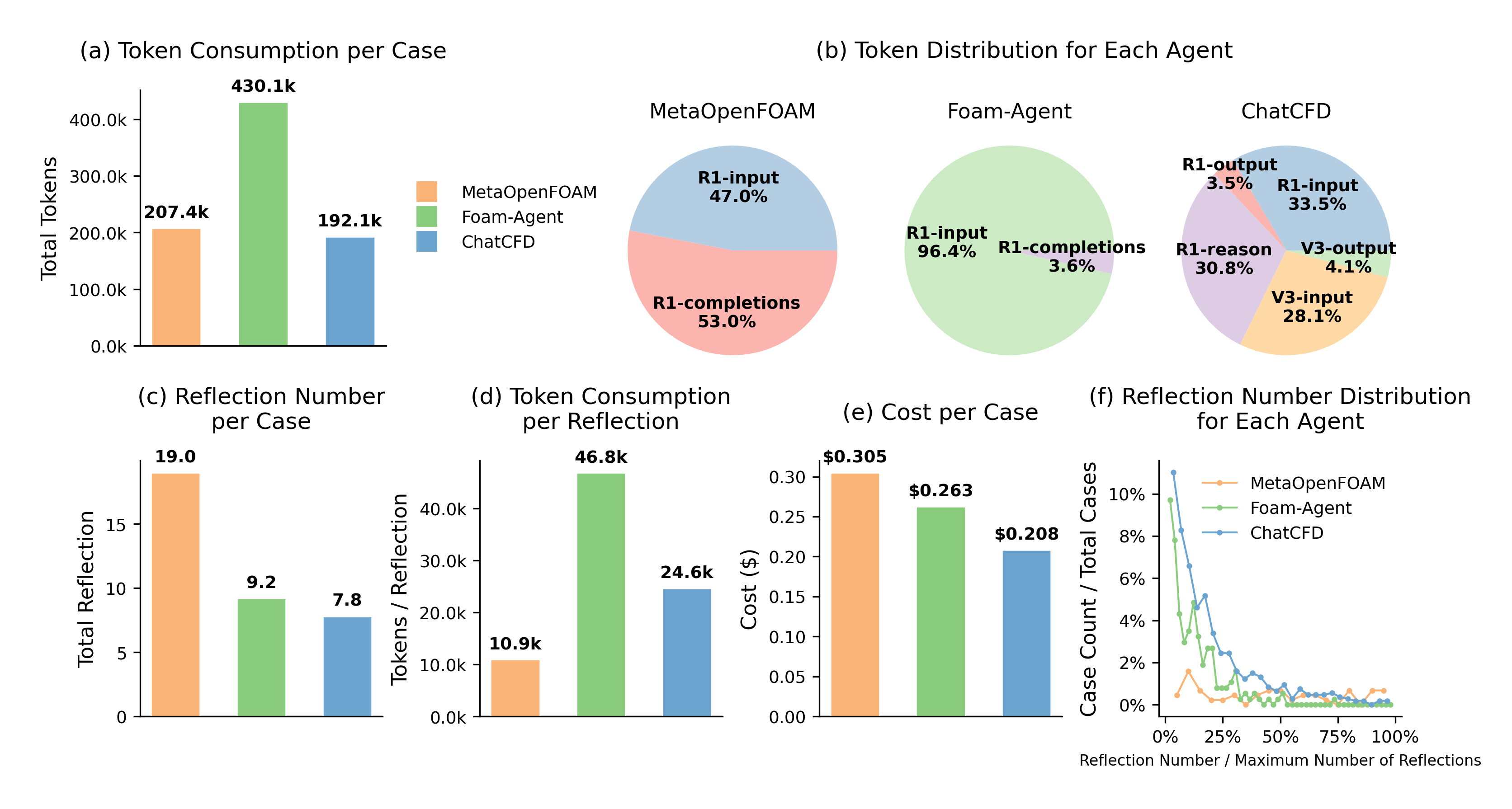}
\vspace{0 pt}
\caption{Performance statistics for different agents across 205 benchmark tutorial cases. (a) Average token consumption per case. (b) Distribution of token consumption. (c) Average number of reflection iterations per case. (d) Average token consumption per reflection iteration. (e) Average computational cost (in monetary terms). (f) Distribution of reflection iteration ratios, excluding zero and limit-reaching cases.}
\label{fig_205_stat}
\end{figure}
The performance of ChatCFD, MetaOpenFOAM, and Foam-Agent was evaluated through detailed metrics on 205 benchmark tutorial cases from OpenFOAM, focusing on token consumption, reflection iterations, and computational cost. Figure \ref{fig_205_stat} presents a comparative analysis of these metrics, highlighting ChatCFD’s efficiency and robustness in automated CFD workflows.
As shown in Figure \ref{fig_205_stat}(a), ChatCFD exhibits the lowest average token consumption per case, approximately half that of Foam-Agent, with MetaOpenFOAM ranking second. This efficiency underscores ChatCFD’s high success rate combined with reduced costs, making it practical for large-scale CFD applications. While ChatCFD and MetaOpenFOAM have comparable total token consumption, Figure \ref{fig_205_stat}(e) reveals a significant cost disparity, with ChatCFD’s computational cost being approximately 1.5 times lower than MetaOpenFOAM’s. This is primarily due to ChatCFD’s strategic use of the cost-efficient \texttt{DeepSeek-V3} model (with a cost half that of \texttt{DeepSeek-R1}) for specific pipeline stages, as illustrated in Figure \ref{fig_205_stat}(b). In contrast, MetaOpenFOAM’s lower success rate necessitates frequent error corrections, generating additional file content and inflating costs.
Figure \ref{fig_205_stat}(c) shows that MetaOpenFOAM requires the highest average number of reflection iterations per case, followed by Foam-Agent, with ChatCFD requiring the fewest. However, Figure \ref{fig_205_stat}(d) indicates that MetaOpenFOAM consumes fewer tokens per reflection iteration due to its simpler error-correction approach, which dilutes the high token cost of initial case setup across multiple iterations. Conversely, ChatCFD and Foam-Agent employ more sophisticated error-reflection strategies. Foam-Agent, for instance, maintains an “error correction trajectory” that logs modifications to file $i$ at iteration $j$, supplemented by contextual data from related files and reference cases. This increases its per-iteration token consumption. ChatCFD, while similarly providing contextual data, optimizes efficiency by maintaining both short-term modification trajectories and long-term reflection summaries. This dual approach enables precise filtering of irrelevant information and concise summarization of critical context, significantly reducing token consumption during LLM interactions compared to Foam-Agent, as evidenced in Figure \ref{fig_205_stat}(d).
Figure \ref{fig_205_stat}(f) depicts the distribution of reflection iteration ratios, defined as the number of reflections divided by the maximum allowed iterations, reflecting each agent’s assessment of case complexity. The x-axis represents the reflection ratio, and the y-axis indicates the proportion of cases at each complexity level. Higher curves indicate stronger error-correction capabilities at a given reflection ratio. All three agents exhibit a rapid decline in solved cases followed by saturation, suggesting that increasing reflection iterations beyond a certain point yields diminishing returns. For Foam-Agent, with a reflection limit of 49 iterations, the curve flattens at approximately 50\% (around 25 iterations), indicating that additional reflections beyond this threshold rarely resolve errors. ChatCFD’s curve consistently lies above the others, demonstrating more effective reflections that are likelier to correct errors. However, even for ChatCFD, reflections beyond 75\% of the limit (approximately 22 iterations) often fail to resolve remaining errors, highlighting a current limitation in LLM-based CFD agents where complex case errors require advanced knowledge integration or alternative strategies beyond iterative reflection.

\subsection{Physical Fidelity Evaluation and Semantic Error Analysis}
\label{sec:phyically_correct}

\textcolor{black}{In Section~\ref{sec:OpenFOAM_tutorials_cases}, we assessed ChatCFD, MetaOpenFOAM, and Foam-Agent using success rate as the primary metric. While this allows rapid quantitative evaluation, successful execution does not ensure physical meaningfulness in CFD. To address this, we introduce physical fidelity as a core metric, evaluating whether executed simulations accurately capture user-specified physics. We also re-examine the link between run success and physical fidelity via agent-generated cases.}

\textcolor{black}{We define physical fidelity through a three-tier protocol. A case qualifies only if it meets all criteria without simplifications that compromise the scientific objective or user requirements:}

\begin{itemize}
    \item \textcolor{black}{\textbf{Conditions Verification}: Initial and boundary conditions match the user-described scenario.}
    \item \textcolor{black}{\textbf{Model Confirmation}: Physical properties, models, and parameters are appropriate and consistent.}
    \item \textcolor{black}{\textbf{Flow Feature Comparison}: Post-processed visualizations align with ground-truth flow characteristics.}
\end{itemize}

\begin{figure}[h!]
  \centering
  \includegraphics[width=0.9\linewidth]{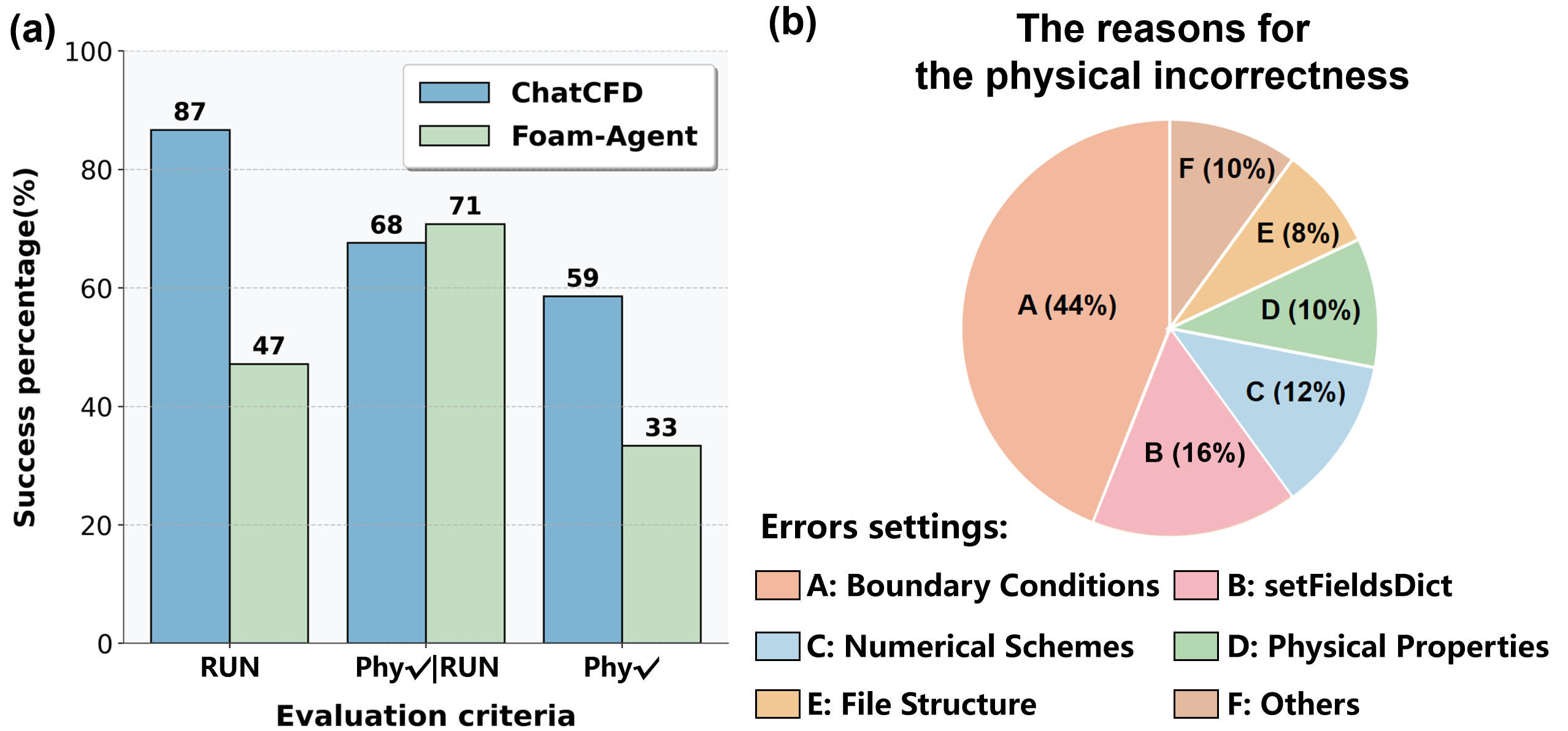}
  \vspace{0 pt}
  \caption{\textcolor{black}{Physical fidelity evaluation of ChatCFD. (a) Success rates of ChatCFD and Foam-Agent under three criteria: run — execution success (simulation completes without crashing); phy$\mid$run — physical fidelity conditioned on execution success (proportion of runnable cases that are physically correct); phy — overall physical fidelity (proportion of all cases that are both runnable and physically meaningful). (b) Root-cause analysis of semantic failures in ChatCFD-generated cases that execute successfully but lack physical fidelity, categorized by error type (boundary/initial conditions, physical properties/models, numerical schemes, etc.).}}
  \label{fig:phy}
\end{figure}

\textcolor{black}{Using this protocol, we analyzed ChatCFD and Foam-Agent results (Fig.~\ref{fig:phy}). ChatCFD's physical fidelity rate among runnable cases (phy|run) is lower than Foam-Agent's, but its overall rate is higher, confirming superior performance. Note that phy|run, limited to runnable cases, may inflate for agents handling only simple setups.}

\textcolor{black}{Additionally, we summarized the causes of physical incorrectness in runnable cases. The results, shown in Fig \ref{fig:phy}, indicate that incorrect settings of boundary conditions, initial conditions, and initial fields in multiphase flows are the main contributors, accounting for 60 \% of the issues, as these require deep physical insight and geometric awareness. Such misconfigurations rarely trigger convergence failures, allowing apparent success that misleads agents. Similarly, omissions in \texttt{setFieldsDict} often evade detection, exacerbating infidelity.}

\textcolor{black}{To further enhance transparency, we integrated a Physics Interpreter, evaluated using DeepSeek-V3 and GPT-4o for summary rationality. It achieved 97.4\% fidelity, confirming accurate natural-language conveyance of the agent's intended physics.}

\textcolor{black}{The contrast between the near-perfect fidelity of the Physical Interpreter (97.4\%) and the overall Physical Fidelity of only 68.12\% is particularly interesting. This suggests that while a seemingly simple user request (e.g., "2-D flow around a cylinder at Re = 100") can be almost perfectly restated in natural language by the LLM, its faithful implementation in executable OpenFOAM code requires dozens of tightly interdependent, domain-specific settings (such as inlet velocity profile, turbulence suppression, kinematic viscosity, blockage ratio, spanwise boundary conditions, etc.). This highlights a key limitation of current LLMs: they excel at high-level linguistic narration but consistently struggle to enforce the complete chain of physical constraints within the generated executable code.}

\begin{CJK*}{UTF8}{gbsn}




\end{CJK*}

\subsection{Ablation Study on ChatCFD Architectural Components}
\label{sec:Ablation_db}

\textcolor{black}{To quantitatively assess the contribution of the proposed database system and functional modules to the overall system performance, we conducted an ablation study using $20\%$ of the cases sampled from the OpenFOAM benchmark test set. It should be noted that the degree of influence exerted by different components will vary due to the inherent differences in task types. Consequently, for the specific task of literature replication, we will perform additional ablation experiments in Section \ref{sec:model_ablation} for further, dedicated analysis.}

\textcolor{black}{The first four experiments were designed to evaluate the impact of the knowledge databases on ChatCFD's performance. This knowledge base primarily influences Stage2 and \texttt{Case File Generation} and \texttt{Error Correction and Reflection} in Stage3. Simultaneously, the experimental groups "w/o mod1" (removal of the Error Locator Module) and "w/o mod2" (removal of the Reflection Module) were used to verify and quantify the efficacy of the error locator and reflection mechanisms, with their influence predominantly focused on \texttt{Error Correction and Reflection} in Stage3.}

\textcolor{black}{The impact of the first four knowledge base ablation experiments on ChatCFD's performance, ranked from highest to lowest influence, is as follows: Solver Template DB (db4, 48 \%), Boundary Condition DB (db2, 57 \%), Parameter Dimension DB (db3, 68 \%), and File Dependency and Structure DB (db1, 73 \%). The results show that the Solver Template DB is utilized across all operational stages of ChatCFD and is crucial for leveraging the Large Language Model's few-shot learning capability to generate initially correct case files. }

\begin{figure}[h!]
  \centering
  \includegraphics[width=0.8\linewidth]{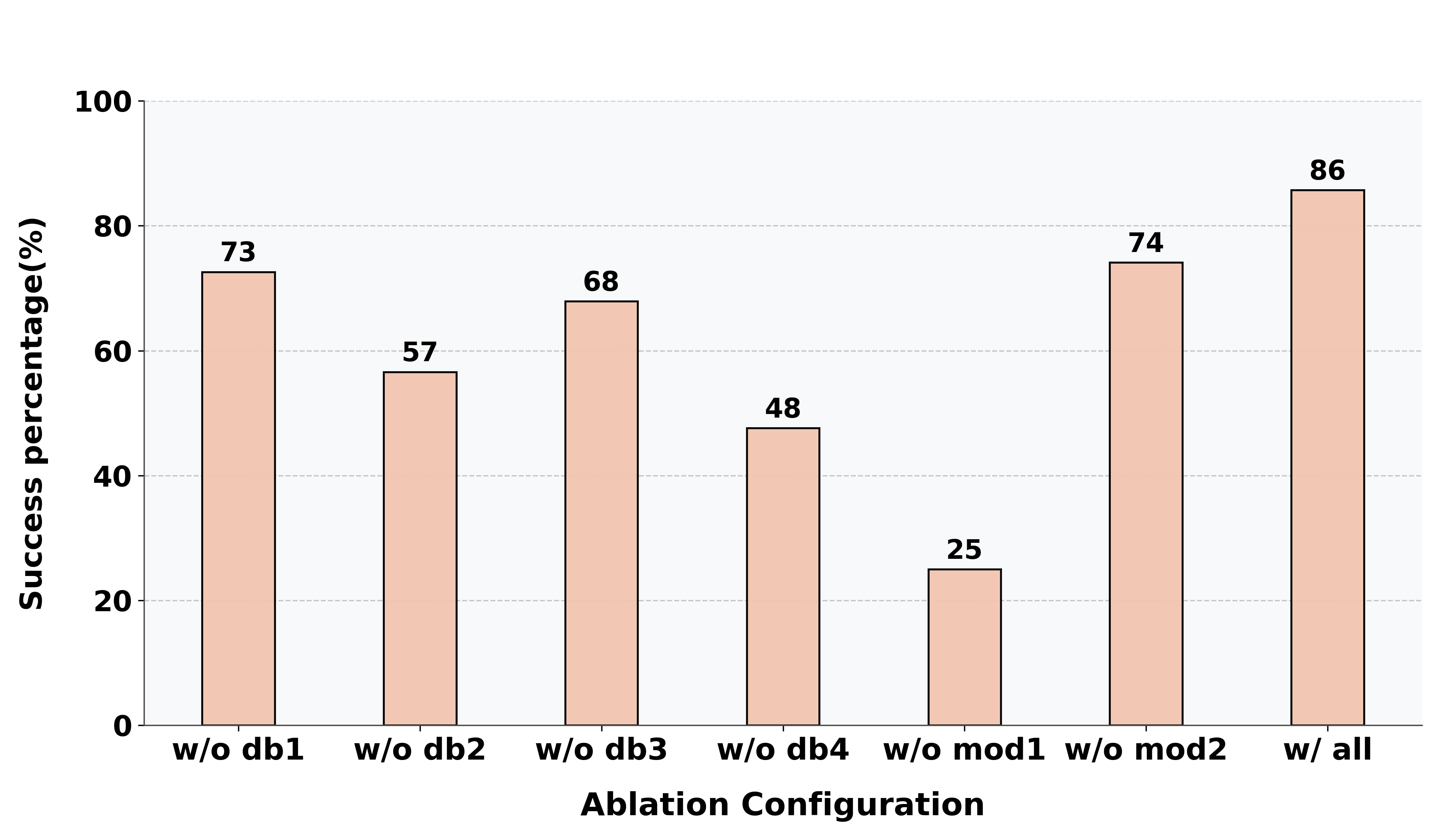}
  \vspace{0 pt}
  \caption{\textcolor{black}{Ablation study quantifying the contribution of individual prior-knowledge components and core modules to ChatCFD’s configuration accuracy.  
- \textbf{w/o File Dependency DB (DB1)}: removes the structured JSON database that enforces mandatory file hierarchy and dictionary dependencies.  
- \textbf{w/o Boundary Condition DB (DB2)}: disables the boundary-condition template database (no validated keyword requirements).  
- \textbf{w/o Parameter Dimension DB (DB3)}: disables automatic dimension enforcement and file-format checks.  
- \textbf{w/o Solver Template DB (DB4)}: removes all reference tutorial snippets and metadata.  
- \textbf{w/o \textcolor{black}{Error Locator Module} (Module 1)}: disables the error-positioning module.  
- \textbf{w/o Reflection (Module 2)}: disables the iterative reflection loop.  
- \textbf{w/ all}: full ChatCFD system (all four prior-knowledge databases + all function modules).}}
  \label{fig:ablation_experiment}
\end{figure}

\textcolor{black}{Fig \ref{fig:ablation_experiment} shows that the removal of the Error Locator Module (w/o mod1) resulted in the lowest success rate. Here, we bypassed it and prompted DeepSeek-R1 to modify files based solely on error messages. The experimental results reveal two findings: 1) Accurate identifying of the erroneous file is a critical step for enhancing the agent's efficiency during the reflection and correction phase (Stage 3); and 2) Relying only on error messages is ineffective. Its shortcomings are particularly pronounced when dealing with errors that do not explicitly point to the source file, such as dimensional errors. Removing the Reflection Module (w/o mod2) causes the smallest performance drop because it primarily benefits high-difficulty cases, which constitute only a minor fraction of the benchmark. In contrast, the other components—knowledge bases and Error Locator—impact cases across all difficulty levels.}




\subsection{\textcolor{black}{Flexibility Evaluation of Alternative OpenFOAM Configurations}}

\textcolor{black}{To evaluate ChatCFD’s ability to adapt to different physical assumptions and numerical settings, we designed two targeted experiments:}

\begin{itemize}
  \item \textcolor{black}{\textbf{Solver selection across regimes} — Given only a textual description, ChatCFD autonomously determines the flow regime (compressible/incompressible, steady/transient) and selects the appropriate solver.}
  \item \textcolor{black}{\textbf{Turbulence-model switching} — With mesh and boundary conditions fixed, only the turbulence closure is changed, requiring automatic rewriting of \texttt{constant/turbulenceProperties} and related dictionaries.}
\end{itemize}

\begin{figure}[h!]
\centering
\includegraphics[width=4in]{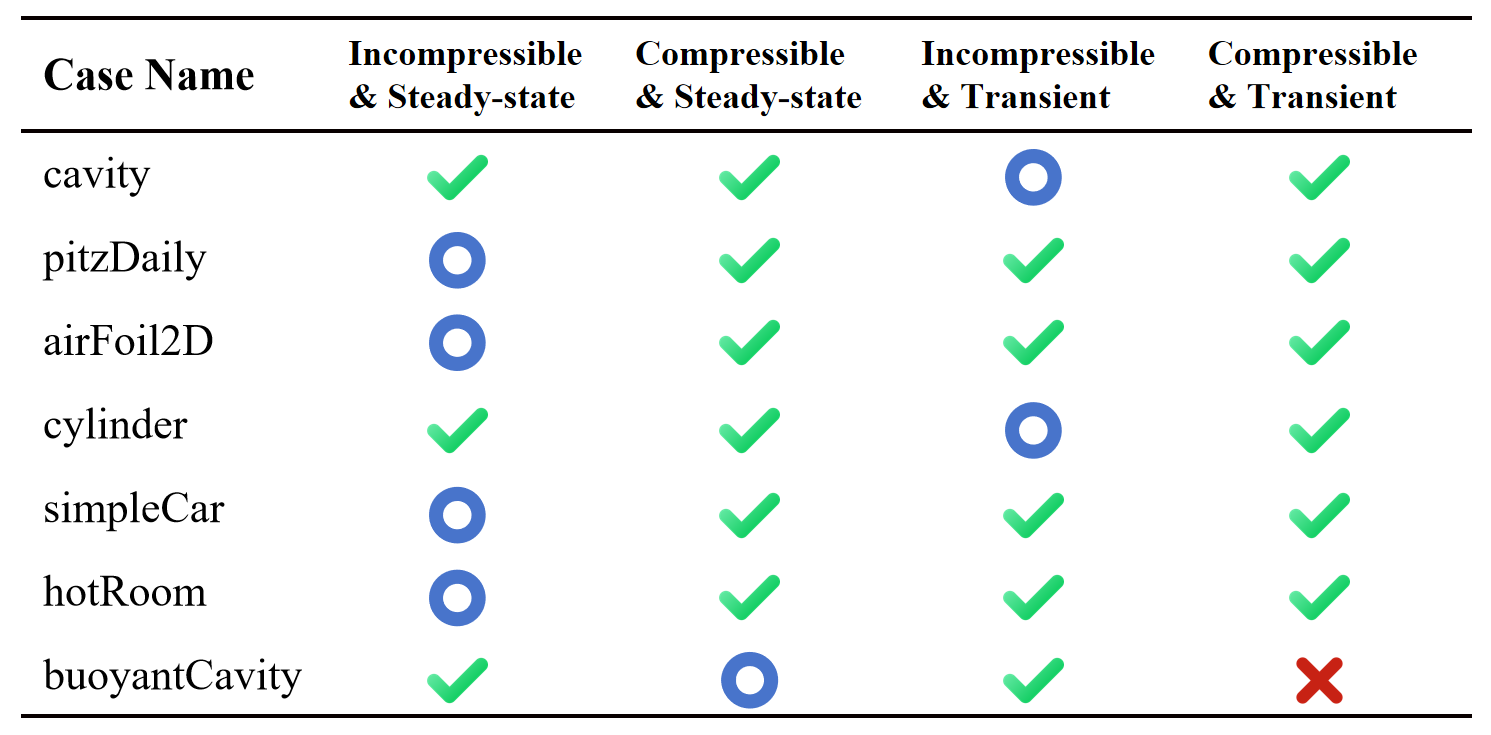}
\vspace{8 pt}
\caption{\textcolor{black}{Solver flexibility under varying physical regimes. Blue circles indicate the reference (tutorial) setting; green checks mark successful execution with the identified solver; red crosses denote failure. ChatCFD correctly selects the appropriate solver in 20/21 of cases, including unseen compressible-transient scenarios.}}
\label{fig_Alter_s}
\end{figure}

\begin{figure}[h!]
\centering
\includegraphics[width=4in]{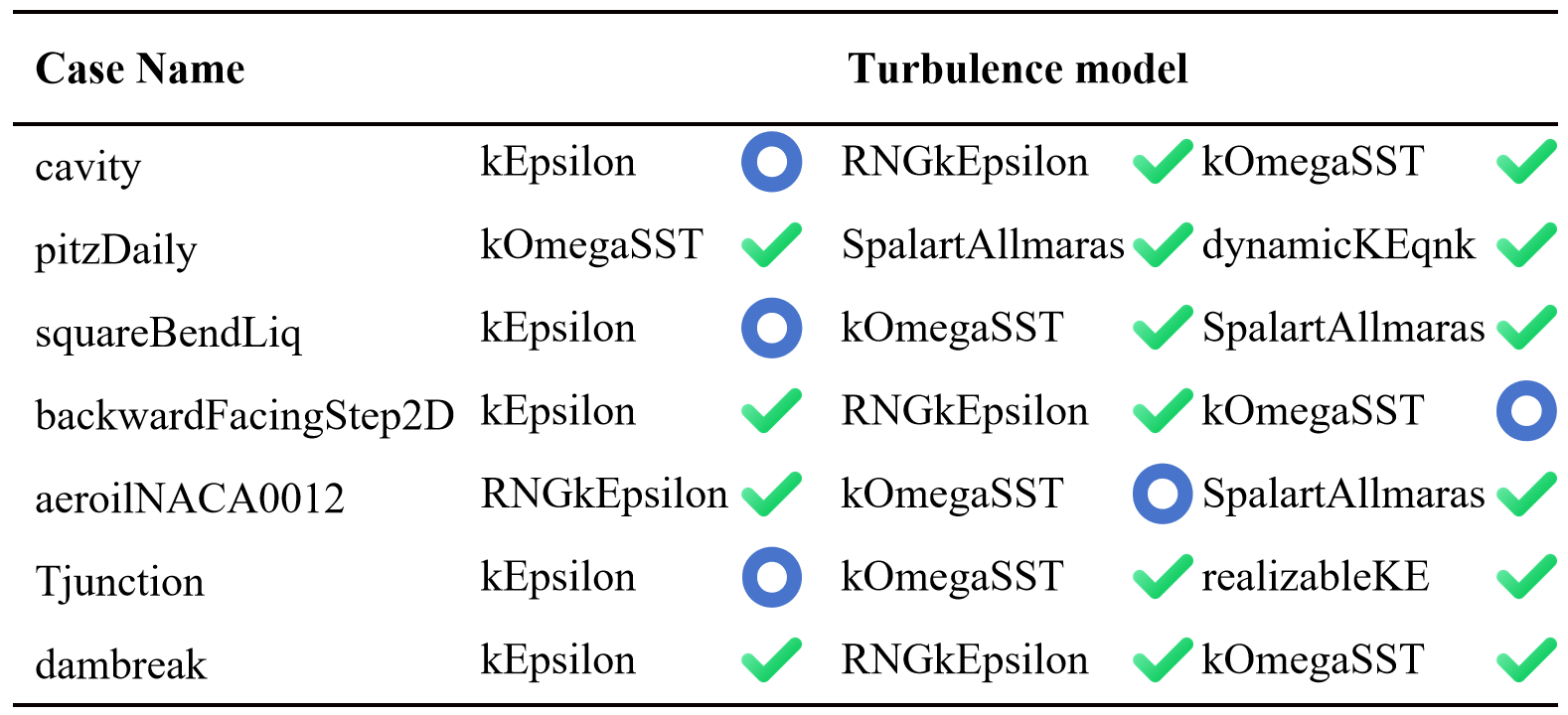}
\vspace{8 pt}
\caption{\textcolor{black}{Turbulence-model flexibility. ChatCFD successfully switches between  k-$\epsilon$, k-$\omega$ SST, and other closures for all tested cases, including the originally laminar damBreak (100\% success). Symbols follow Figure~\ref{fig_Alter_s}.}}
\label{fig_Alter_t}
\end{figure}

\textcolor{black}{Fig \ref{fig_Alter_s} summarizes the results of Experiment 1. ChatCFD correctly identified the required solver in 20 of 21 cases, including the unseen ``simpleCar'' compressible-transient configuration (not present in the database). The sole failure occurred on ``buoyantCavity'' under compressible-transient assumptions, which currently exceeds ChatCFD’s supported scope.
In the turbulence-switching experiment (Figure~\ref{fig_Alter_t}), ChatCFD successfully applied k-$\epsilon$, k-$\omega$ SST, and laminar closures to every case—including the originally laminar \texttt{damBreak}—demonstrating robust handling of non-default configurations.
These results confirm ChatCFD’s strong generalization: it is not limited to tutorial-default settings but can reason about physically valid alternatives, significantly enhancing its practical utility for practical CFD exploration.
}

\subsection{Evaluation of CFD Cases Reproduced from Literature}
\label{sec:CFD_cases}

\begin{figure}[h!]
\centering
\includegraphics[width=5in]{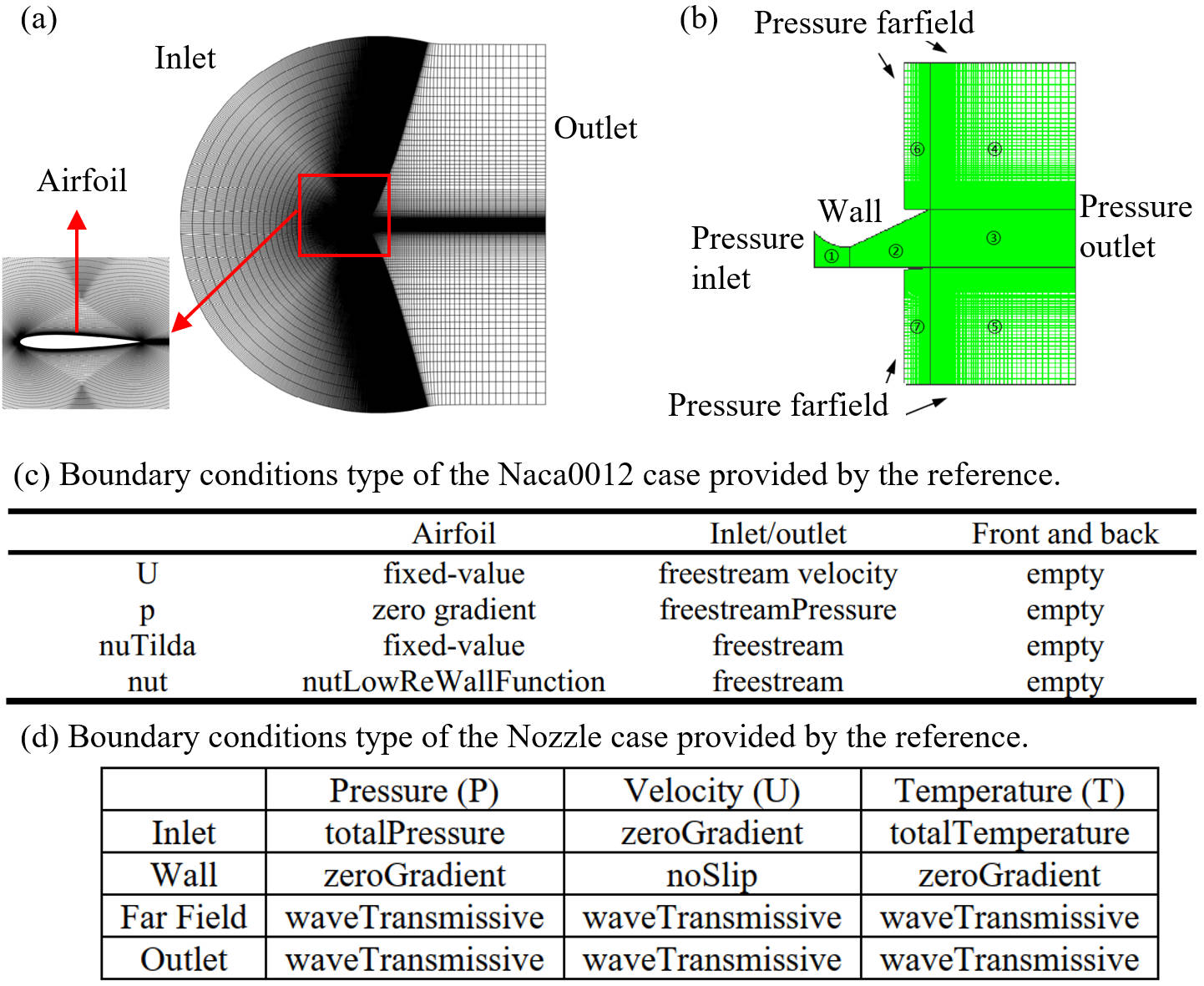}
\vspace{8 pt}
\caption{Visualization of two literature-derived CFD cases. (a) Computational mesh for the NACA0012 airfoil case \cite{sun2023comparison}. (b) Computational mesh for the Nozzle case \cite{yu2023comparative}. (c) Boundary condition types for the NACA0012 case. (d) Boundary condition types for the Nozzle case.}
\label{fig_two_cases_with_bcs}
\end{figure}

To evaluate ChatCFD’s ability to handle complex, real-world scenarios, two representative literature-derived cases were selected: an incompressible flow case and a compressible flow case, as illustrated in Figure \ref{fig_two_cases_with_bcs}. These cases were chosen for their comprehensive documentation in the source literature, providing detailed specifications for solver configurations, turbulence models, and mesh characteristics critical for accurate CFD reproduction. The incompressible flow case, based on Sun et al. \cite{sun2023comparison}, involves a NACA0012 airfoil at a 10° angle of attack, simulated using the \texttt{simpleFoam} solver with the Spalart-Allmaras turbulence model. The compressible flow case, drawn from Yu et al. \cite{yu2023comparative}, models a nozzle with a pressure ratio of 3, employing the \texttt{rhoCentralFoam} solver and the Spalart-Allmaras model.

Given reported challenges in modifying OpenFOAM turbulence models using large language models (LLMs), as noted by Pandey et al. \cite{pandey2025openfoamgpt}, additional experiments were conducted to explore robustness across various turbulence models (e.g., k-$\epsilon$, k-$\omega$ SST). Table \ref{tab:chatcfd_test_cases} provides a comprehensive summary of all case configurations and experimental iterations. High-fidelity computational meshes were generated to ensure accuracy. The experimental protocol was structured as follows: Cases 1 (NACA0012) and 5 (nozzle) underwent an extensive model ablation study, with 10 iterations per case across five system configurations to assess sensitivity to model variations. Cases 2 through 4, using different turbulence models, were evaluated based on the full ChatCFD system configuration, with 10 iterations each, to focus on performance under optimal settings. This rigorous methodology resulted in 130 experimental runs, forming a robust dataset for analyzing ChatCFD’s performance in complex, literature-derived scenarios. The results, detailed in subsequent analyses, demonstrate ChatCFD’s capability to adapt to sophisticated CFD configurations, complementing its strong performance on benchmark tutorial and perturbed variant cases.

\begin{table}[htbp]
\caption{Summary of Test Cases for ChatCFD Validation}
\centering
\begin{tabular}{@{}llll@{}}
\toprule
\textbf{Case ID} & \textbf{Case Mesh} & \textbf{Physical Models} & \textbf{Experimental Design} \\
\midrule
Case 1 & NACA0012 & \begin{tabular}[t]{@{}l@{}}\texttt{simpleFoam} with\\ Spalart-Allmaras\end{tabular} & \begin{tabular}[t]{@{}l@{}}10 runs $\times$ 5 system\\ configurations\end{tabular} \\
\addlinespace
Case 2 & NACA0012 & \begin{tabular}[t]{@{}l@{}}\texttt{simpleFoam} with\\ k-$\omega$ SST\end{tabular} & \begin{tabular}[t]{@{}l@{}}10 runs $\times$ the complete \\ system configuration\end{tabular} \\
\addlinespace
Case 3 & NACA0012 & \begin{tabular}[t]{@{}l@{}}\texttt{simpleFoam} with\\ k-$\epsilon$\end{tabular} & \begin{tabular}[t]{@{}l@{}}10 runs $\times$ the complete\\ system configuration\end{tabular} \\
\addlinespace
Case 4 & NACA0012 & \begin{tabular}[t]{@{}l@{}}\texttt{simpleFoam} with\\ RNGk-$\epsilon$\end{tabular} & \begin{tabular}[t]{@{}l@{}}10 runs $\times$ the complete\\ system configuration\end{tabular} \\
\addlinespace
Case 5 & Nozzle & \begin{tabular}[t]{@{}l@{}}\texttt{rhoCentralFoam} with\\ hePsiThermo and\\ Spalart-Allmaras\end{tabular} & \begin{tabular}[t]{@{}l@{}}10 runs $\times$ 5 system\\ configurations\end{tabular} \\
\bottomrule
\end{tabular}
\label{tab:chatcfd_test_cases}
\end{table}

\subsubsection{Ablation Study on ChatCFD Configurations for Literature Reproduction} 
\label{sec:model_ablation}
To systematically dissect the contributions of ChatCFD’s core components, ablation experiments were conducted across five distinct system configurations. These configurations vary in the complexity of case file generation (Stage 2, Section \ref{sec:initiating_case_files}), the error-handling logic, and the integration of retrieval modules (\texttt{ReferenceRetriever} and \texttt{ContextRetriever}) in Stage 3 (Correcting Case Files). The configurations are defined as follows:

\begin{itemize}[leftmargin=*]
    \item \textbf{Configuration A (Baseline System)}: Employs a simplified parameter extraction process in Stage 2, bypassing the three-step hierarchical extraction. In Stage 3, error correction is limited to a basic 'general error' pathway without \texttt{ReferenceRetriever} (tutorial-based retrieval).
    \item \textbf{Configuration B (Baseline with \texttt{ReferenceRetriever}) and Simplified Setup Extraction}: Retains the simplified Stage 2 extraction but enhances Stage 3 with the \texttt{ReferenceRetriever} module, enabling tutorial-based error correction within the general error pathway.
    \item \textbf{Configuration C (Enhanced Setup and \texttt{Reference Retrieval})}: Implements the full three-step setup extraction and configuration validation in Stage 2. Stage 3 includes general error correction augmented by the \texttt{ReferenceRetriever} module for improved accuracy.
    \item \textbf{Configuration D (Full Error Reflection with Simplified Extraction)}: Uses the simplified Stage 2 extraction but incorporates both general and persistent error correction modules in Stage 3, supported by \texttt{ReferenceRetriever} and \texttt{ContextRetriever} for comprehensive error handling.
    \item \textbf{Configuration E (Complete System)}: Represents the full ChatCFD pipeline, featuring comprehensive case setup extraction and validation in Stage 2 and complete error reflection modules in Stage 3, integrating both \texttt{ReferenceRetriever} and \texttt{ContextRetriever} for optimal performance.
\end{itemize}

\begin{figure}[h!]
\centering
\includegraphics[width=5in]{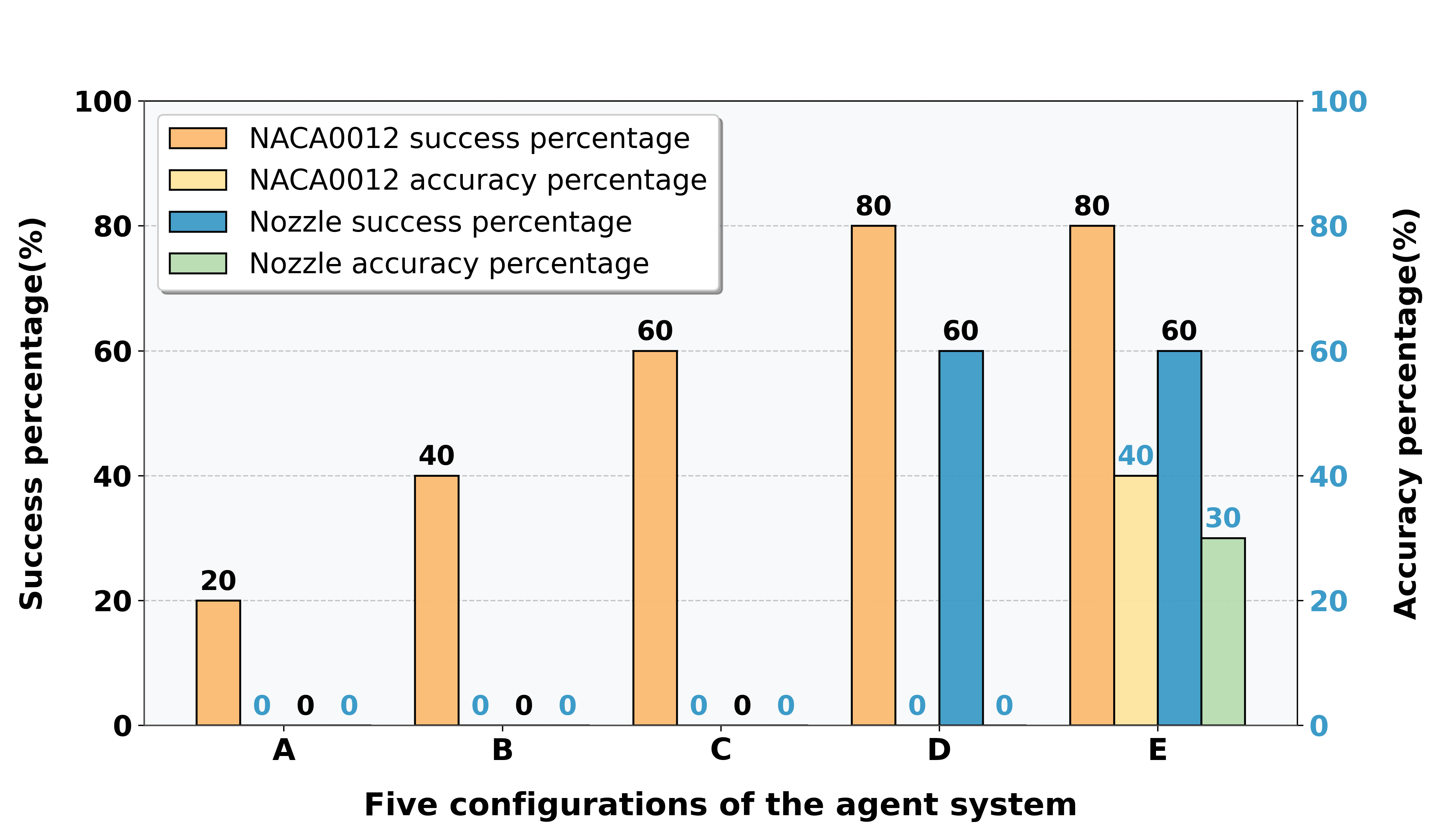}
\vspace{8 pt}
\caption{Success rates of ChatCFD across five system configurations (A--E) for literature-derived cases: NACA0012 (Case 1) and Nozzle (Case 5). The horizontal axis represents configurations, and the vertical axis quantifies success rates. The 10-step operational success rate indicates completion of 10 simulation steps (time steps for transient cases or iterations for steady-state cases) without critical errors. The accurate configuration success rate reflects precise adherence to source article specifications.}
\label{fig_model_fusion_analyses}
\end{figure}

Figure \ref{fig_model_fusion_analyses} illustrates ChatCFD’s performance across Configurations A through E for two literature-derived cases: NACA0012 (Case 1) and Nozzle (Case 5). Two metrics were evaluated: (1) the \textit{10-step operational success rate}, defined as the ability to complete 10 simulation steps (time steps for transient cases or iterations for steady-state cases) without critical errors, and (2) the \textit{accurate configuration success rate}, indicating exact compliance with the solver, turbulence model, and boundary condition specifications in the source articles. The 10-step criterion enables rapid assessment of operational stability, particularly for the reflection mechanism, as literature often lacks specified simulation durations. Configurations achieving this milestone were generally free of major operational faults, though minor deviations in boundary condition parameters could persist. Cases meeting the accurate configuration criterion exhibited simulation results in strong agreement with published data, as detailed in the appendix. These results underscore ChatCFD’s robustness in handling complex, literature-derived configurations, complementing its performance on benchmark tutorial and perturbed variant cases.

For the NACA0012 Case 1, the data in Figure \ref{fig_model_fusion_analyses} show a clear progression in the 10-step operational success rate with increasing system sophistication. The baseline Configuration A achieved a 20\% success rate. This trend highlights the positive impact of incrementally integrating retrieval modules and OpenFOAM-specific knowledge. However, despite its high operational success, Configuration D failed to achieve any accurate configuration success, unable to precisely replicate the case setup from the source article. This crucial gap was addressed by Configuration E, which incorporates comprehensive article interpretation and setup extraction as part of its advanced Stage 2 processing. With Configuration E, the accurate configuration success rate for the NACA0012 case improved significantly to 40\%, while maintaining the 80\% operational success rate. Analysis of the \texttt{DeepSeek-R1} model’s internal reasoning revealed that, without robust article interpretation and strict compliance with documented specifications in Stage 2, the LLM often defaults to generic boundary conditions and simplified parameters, such as standard \texttt{inlet}-type conditions for inflow velocities or \texttt{pressureOutlet}-type conditions for outflow pressures, instead of the specialized \texttt{freestream} conditions required for the NACA0012 airfoil. These erroneous configurations underscore the critical role of comprehensive case setup extraction and validation in Stage 2, which Configuration E leverages to ensure fidelity to complex CFD requirements, enhancing ChatCFD’s performance in complicated scenarios.

For the Nozzle Case 5, Figure \ref{fig_model_fusion_analyses} demonstrates a significant enhancement in the 10-step operational success rate with the integration of the \texttt{ContextRetriever} module. Configurations A through C, lacking \texttt{ContextRetriever}, achieved a 0\% success rate for this metric, defined as completing 10 simulation steps (time steps for transient cases or iterations for steady-state cases) without critical errors. Configuration D, incorporating \texttt{ContextRetriever}, improved this rate to 60\%. This advancement stems from \texttt{ContextRetriever}’s ability to diagnose complex coupling-related errors within the MCP framework, addressing a key limitation of earlier configurations. For example, errors in thermophysical parameters, such as compressibility ($\Psi$), were often misattributed to the \texttt{constant/thermophysicalProperties} file in prior configurations. In reality, these errors frequently arise from inconsistencies in equation of state specifications within boundary condition entries in field files (e.g., \texttt{0/p}, \texttt{0/T}). By analyzing all relevant directories (\texttt{0/}, \texttt{constant/}, \texttt{system/}), \texttt{ContextRetriever} enables precise error locator and effective correction, enhancing performance for complex cases.

The differential impact of \texttt{ContextRetriever} between the NACA0012 Case 1 and Nozzle Case 5 reflects their distinct flow physics. The NACA0012 case, an incompressible flow simulation using the \texttt{pisoFoam} solver, exhibits weak pressure-velocity coupling, reducing the need for advanced error handling. Conversely, Nozzle Case 5, a compressible flow scenario with the \texttt{rhoCentralFoam} solver, involves strong interdependencies among density, temperature, and pressure, governed by the equation of state. The \texttt{ContextRetriever} module’s advanced analysis of inter-file dependencies proves particularly effective for the Nozzle case’s coupled physics, yielding greater performance gains compared to the less coupled NACA0012 case. This highlights \texttt{ContextRetriever}’s proficiency in managing complex physical interactions across diverse CFD scenarios.

Advancing to Configuration E, which integrates comprehensive article interpretation in Stage 2, maintained the 60\% 10-step operational success rate for Nozzle Case 5 while increasing the accurate configuration success rate from 0\% to 30\%. This metric reflects precise adherence to source specifications. Consistent with findings for the NACA0012 case, Configuration E’s enhanced setup extraction and validation ensure greater configuration fidelity across flow regimes, reinforcing ChatCFD’s robustness in literature-derived cases. 

\begin{figure}[h!]
\centering
\includegraphics[width=6.25in]{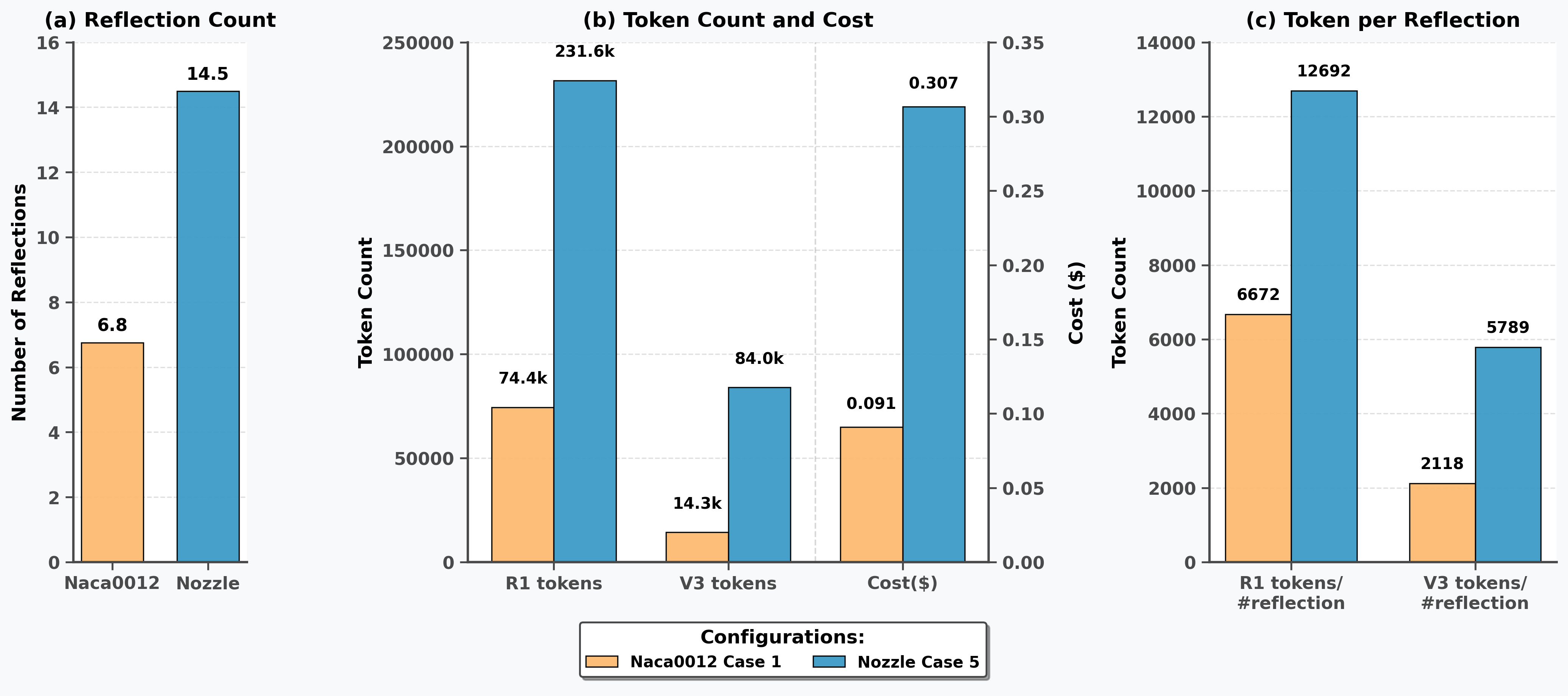}
\caption{ChatCFD’s performance metrics for accurate configuration of literature-derived NACA0012 Case 1 and Nozzle Case 5. Metrics include:  number of reflection iterations (\#reflection), total token consumption for \texttt{DeepSeek-R1} and \texttt{DeepSeek-V3} models (R1/V3 total token), execution cost (Cost \$), and average token consumption per reflection round for both models (R1/V3 token/\#reflection).}
\label{fig_physical_coupling}
\end{figure}

The performance disparity between the literature-derived NACA0012 Case 1 and Nozzle Case 5 arises from two key factors: (1) the increased complexity of managing numerous files with intricate content in compressible flows, and (2) the stringent requirement for consistency across coupled thermophysical models (e.g., \texttt{0/p}, \texttt{0/T}, \texttt{constant/thermophysicalProperties}). The Nozzle case, a compressible flow simulation using \texttt{rhoCentralFoam}, involves strong coupling among pressure, density, temperature, and the equation of state, necessitating precise dimensional and physical alignment across governing equations and material properties. This contrasts with the NACA0012 case, an incompressible flow simulation using \texttt{pisoFoam} with weaker pressure-velocity coupling, which imposes fewer constraints.
Additionally, an implicit bias in LLM training data, predominantly focused on incompressible flow scenarios, leads to erroneous defaults in compressible cases. For instance, in OpenFOAM, incompressible flows use kinematic pressure ($p_k$, dimensions $L^2T^{-2}$)—dynamic pressure divided by density—while compressible flows require absolute thermodynamic pressure ($p$, dimensions $ML^{-1}T^{-2}$). The \texttt{DeepSeek-R1} model’s tendency to apply incompressible conventions to compressible cases introduces dimensional errors, such as incorrect pressure settings, requiring additional reflection iterations for correction.

Figure \ref{fig_physical_coupling} quantifies these challenges: the Nozzle case requires over twice the reflection iterations, triple the \texttt{DeepSeek-R1} tokens, and six times the \texttt{DeepSeek-V3} tokens compared to the NACA0012 case, resulting in approximately triple the execution cost. The disproportionate \texttt{DeepSeek-V3} token increase (6x vs. 3x for \texttt{DeepSeek-R1}) reflects frequent dimensional inconsistencies and persistent errors, necessitating more invocations of the costlier \texttt{DeepSeek-V3} model. Figure \ref{fig_physical_coupling}(c) confirms this, showing a \texttt{DeepSeek-V3} token multiplier of ~2.7x per reflection for the Nozzle case, compared to ~2x for \texttt{DeepSeek-R1}, relative to the NACA0012 case. These metrics highlight the computational burden of complex flow simulations and underscore ChatCFD’s ability through advanced error correction.

\subsubsection{Influence of Turbulence Models on ChatCFD Efficacy}
\label{sec:turbulence_model}
\begin{figure}[h!]
    \centering
    \includegraphics[width=6.25in]{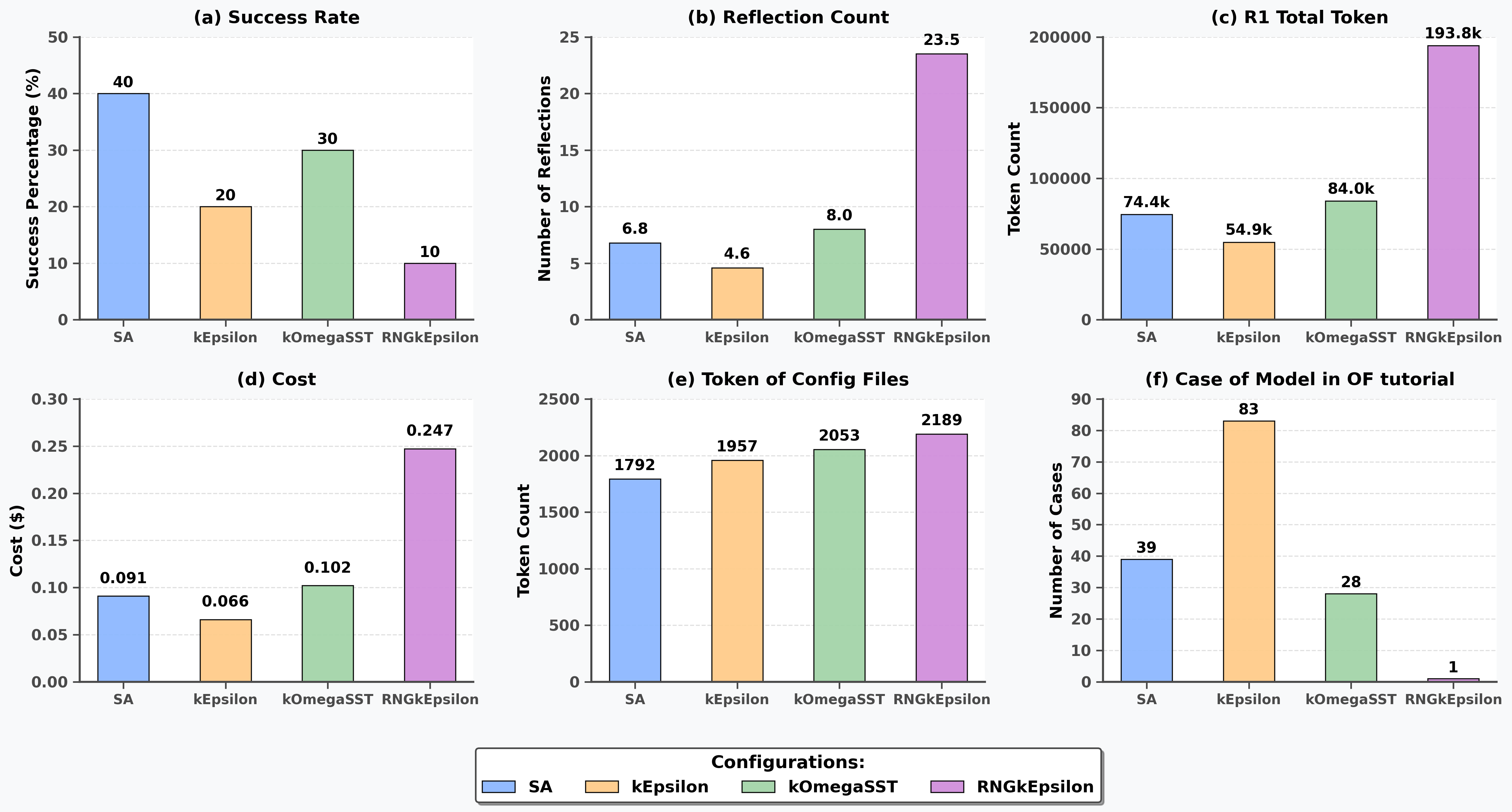}
    \caption{ChatCFD's average performance for accurate configuration of NACA0012 Cases 1 to 4 across turbulence models (Spalart-Allmaras, k-$\epsilon$, k-$\omega$ SST, RNG k-$\epsilon$). Metrics shown include: accuracy (success rate of precise configuration according to literature), number of reflection iterations (\#reflection), total token consumption of the \texttt{DeepSeek-R1} model (R1 total token), execution cost (Cost \$), token count of configuration files (\#token of config files), and the number of cases incorporating each turbulence model in OpenFOAM tutorials (\#Case of model in OF tutorial).}
    \label{fig_four_turbulence_model}
\end{figure}

Figure \ref{fig_four_turbulence_model} presents ChatCFD's performance across four turbulence models—Spalart-Allmaras (SA), k-$\epsilon$, k-$\omega$ SST, and RNG k-$\epsilon$—applied to the literature-derived NACA0012 cases (Cases 1 to 4). The analysis focuses on simulations achieving accurate configuration per published specifications, excluding those with only basic operational success (e.g., 10-step operational success rate). All results were obtained using the complete ChatCFD system (Configuration E).

As shown in Figure \ref{fig_four_turbulence_model}(a), success rates for accurate configuration vary significantly: SA at 40\%, k-$\omega$ SST at 30\%, k-$\epsilon$ at 20\%, and RNG k-$\epsilon$ at 10\%. This demonstrates ChatCFD's ability to handle diverse turbulence models, addressing LLM challenges noted in OpenFOAMGPT \cite{pandey2025openfoamgpt} through structured knowledge integration and RAG.
Performance metrics in Figure \ref{fig_four_turbulence_model}(a,b) reveal consistent profiles for SA, k-$\epsilon$, and k-$\omega$ SST, with an average of ~6 reflection iterations, comparable \texttt{DeepSeek-R1} token usage, and costs. In contrast, RNG k-$\epsilon$ requires ~23.5 reflections—nearly 4x higher—elevating token consumption and costs.

Figure \ref{fig_four_turbulence_model}(c) attributes these disparities to configuration file token counts and model prevalence in OpenFOAM tutorials. While file sizes differ modestly (~1,800 tokens for SA vs. ~2,000 for others), tutorial distribution is more influential: SA, k-$\epsilon$, and k-$\omega$ SST appear in 28+ tutorials each, versus one for RNG k-$\epsilon$.
RNG k-$\epsilon$'s performance deficit stems from: (1) its rarity in engineering applications, leading to sparse LLM training data and poor zero-shot configuration; and (2) limited \texttt{ReferenceRetriever} efficacy due to few OpenFOAM tutorials, hindering guidance for parameters in \texttt{constant/turbulenceProperties}, \texttt{system/fvSchemes}, and \texttt{system/fvOption}. This necessitates greater reliance on the LLM's limited intrinsic knowledge, reducing performance relative to common models.

\subsection{Future Potential of ChatCFD in Multi-Agent CFD Workflows}

Building on the insights from ChatCFD’s performance in benchmark tutorial, perturbed variant, and literature-derived cases, ongoing work explores its potential as a foundational component in advanced engineering workflows. Beyond functioning as a standalone tool for CFD automation, ChatCFD is being adapted through the Model Context Protocol (MCP), a standardized framework for integrating specialized AI agents into a cohesive multi-agent system. MCP facilitates seamless communication and data exchange among agents by defining structured protocols for context sharing, enabling ChatCFD to collaborate with other intelligent tools to address complex, interdisciplinary engineering tasks.

\begin{figure}[h!]
    \centering
    \includegraphics[width=6.25in]{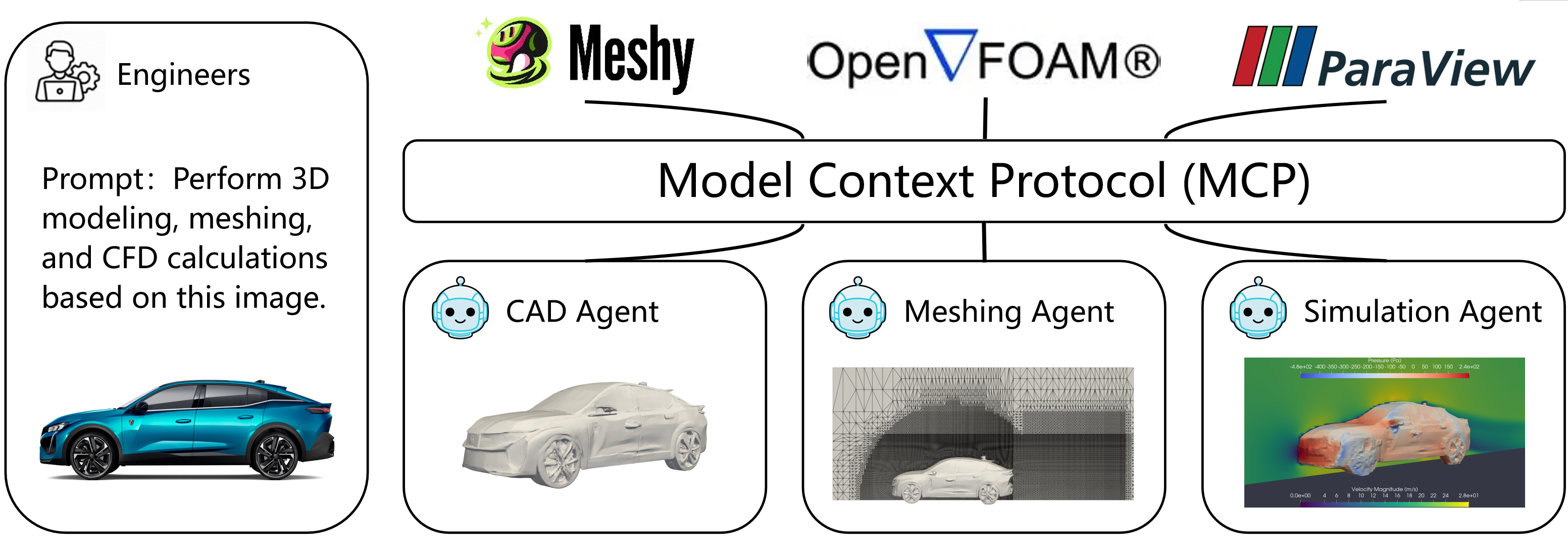}
    \caption{ChatCFD integrated within a multi-agent workflow via the Model Context Protocol (MCP), enabling transformation from a single sentence or image to a complete CFD simulation.}
    \label{fig_mcp}
\end{figure}

As illustrated in Figure \ref{fig_mcp}, this multi-agent workflow enables an end-to-end design cycle for CFD applications. The process begins with a 3D modeling agent that generates a three-dimensional geometry from a natural language description or a simple image, such as a sketch or photograph. A subsequent meshing agent automatically discretizes the geometry into a high-fidelity computational mesh, leveraging tools like commercial mesh generators or open-source alternatives (e.g., \texttt{blockMesh}, \texttt{snappyHexMesh}). This mesh is then passed to the ChatCFD agent, which autonomously configures solver parameters, boundary conditions, and turbulence models (e.g., \texttt{simpleFoam} with Spalart-Allmaras or \texttt{rhoCentralFoam} with k-$\omega$ SST) to execute the CFD simulation, computing key performance metrics such as drag coefficients or pressure distributions. Finally, an optimization agent analyzes the simulation results, providing insights for iterative design refinement, such as adjusting geometry or boundary conditions to minimize drag or enhance flow stability. This MCP agent is available to use and can be found online (https://www.bohrium.com/apps/designagent0812).

This integrated approach highlights ChatCFD’s transformative potential beyond simulation automation. By serving as a standardized computational module within a collaborative multi-agent ecosystem, ChatCFD can contribute to broader engineering design processes, from conceptual design to performance optimization. The MCP framework ensures interoperability, allowing ChatCFD to adapt to diverse tasks, such as aerodynamics, heat transfer, or multiphase flow simulations, by interfacing with agents specialized in geometry generation, mesh refinement, or post-processing. This vision positions ChatCFD as a cornerstone for future AI-driven engineering workflows, paving the way for scalable, interdisciplinary applications in scientific discovery and industrial innovation.

\section{Conclusion}
\label{sec:conclusion}

\textcolor{black}{This paper introduces ChatCFD, an innovative LLM-based CFD agent system for automating OpenFOAM simulations. By integrating a multi-agent architecture, domain-specific structured knowledge bases, precise error locator, and iterative reflection, ChatCFD dramatically outperforms prior systems while exposing fundamental limitations of current LLMs in scientific computing. }

\textcolor{black}{On 315 OpenFOAM benchmark cases, ChatCFD attains an \textbf{82.1\% execution success rate}—far surpassing MetaOpenFOAM (6.2\%) and Foam-Agent (42.3\%). More importantly, we introduced \textbf{physical fidelity} as a rigorous new metric: among all attempts, ChatCFD generates \textbf{physically correct solutions in 68.12\%} of cases, with 60\% of semantic failures traced to subtle boundary/initial condition errors that escape convergence checks. A dedicated \textbf{Physics Interpreter} further achieves \textbf{97.4\% summary fidelity}, revealing the striking gap between LLMs’ fluency in high-level narration and their struggle to enforce dozens of tightly coupled domain constraints in executable code.}

\textcolor{black}{Resource-efficiency analysis on 205 benchmark cases further underscores ChatCFD’s practical superiority. It consumes only \textbf{192.1k tokens and \$0.208 per case}—roughly half the tokens of Foam-Agent and 1.5× cheaper than MetaOpenFOAM—while requiring the fewest reflection iterations on average. This efficiency gain over naive LLM prompting stems from strategic model routing (DeepSeek-V3 for generation, DeepSeek-R1 for reasoning) and a reflection memory that filters irrelevant context. Even under a strict iteration limit, ChatCFD’s reflection trajectory consistently outperforms baselines, solving cases that saturate the others at ~50\% success after 25 iterations. These results demonstrate that structured domain integration not only boosts accuracy but dramatically reduces the operational cost of LLM-driven scientific computing, making high-fidelity CFD automation viable at scale.}

\textcolor{black}{Ablation studies confirm that structured OpenFOAM knowledge is indispensable: removing the Solver Template DB alone collapses accuracy to 48\%, while the Error Locator Module proves the single most critical component for real-world robustness. Flexibility experiments demonstrate ChatCFD’s ability to autonomously select appropriate solvers across compressible/incompressible and steady/transient regimes (95.23\% success) and switch turbulence closures (100\% success), even on unseen configurations.}

\textcolor{black}{By reproducing complex literature cases—NACA0012 airfoil, supersonic nozzles—ChatCFD shows 60–80\% end-to-end success where baselines fail entirely. These results establish potentially \textbf{new benchmarks} for AI-driven CFD and provide the quantitatively grounded evidence of where current LLMs break in scientific tool use.}

ChatCFD effectively addresses the critical gaps in CFD automation agents. First, it integrates domain-specific knowledge through structured RAG and expert-designed agents, mitigating LLM training scarcity to enable precise boundary and initial condition specifications. Second, its modular, MCP-compatible design supports collaborative multi-agent networks, integrating specialized meshing agents to handle complex geometries beyond \texttt{blockMesh}. Third, by leveraging extensive literature-derived corpora with detailed specifications and benchmarks, it will facilitate scalable testing and iterative agent evolution, harnessing millions of CFD papers for refinement and validation in the future. \textcolor{black}{These advances will enable scientists across disciplines to explore bold ideas using CFD that were previously stalled by technical details. By liberating researchers from low-level case-setup burdens, it enables non-experts to perform computational fluid dynamics studies, allows rapid exploration of hypothesis and design spaces, and frees CFD specialists to focus on genuine scientific questions.}

\section*{ACKNOWLEDGEMENTS}
This work is sponsored by the National Natural Science Foundation of China, Grant No. 92470127, and the Overseas Postdoctoral Talents Program in Guangdong Province.

\section*{AUTHOR DECLARATIONS}
\subsection*{Conflict of Interest}

The authors declare no conflicts of interest.

\subsection*{Author Contributions}

\textbf{E Fan}: Methodology, Data curation, Formal analysis, Writing - original draft. 
\textbf{Kang Hu}: Methodology, Software, Data curation, Formal analysis, Writing - original draft. 
\textbf{Zhuowen Wu}: Methodology. 
\textbf{Jiangyang Ge}: Data curation, Formal analysis.
\textbf{Jiawei Miao}: Software, Resources. 
\textbf{Yuzhi Zhang}: Conceptualization, Software, Resources. 
\textbf{He Sun}: Resources, Writing - Review. 
\textbf{Weizong Wang}: Resources, Writing - Review, Funding acquisition.
\textbf{Tianhan Zhang}: Conceptualization, Methodology, Formal analysis, Resources, Writing - Review \& Editing, Visualization, Supervision, Funding acquisition

\subsection*{Data Availability}
The data supporting the findings are available at \url{https://github.com/ConMoo/ChatCFD}.

\bibliographystyle{unsrt}  
\bibliography{references} 
\appendix
\section{Appendix: detailed analysis of two literature-derived cases}
\begin{figure}[h!]
\centering
\includegraphics[width=6.25in]{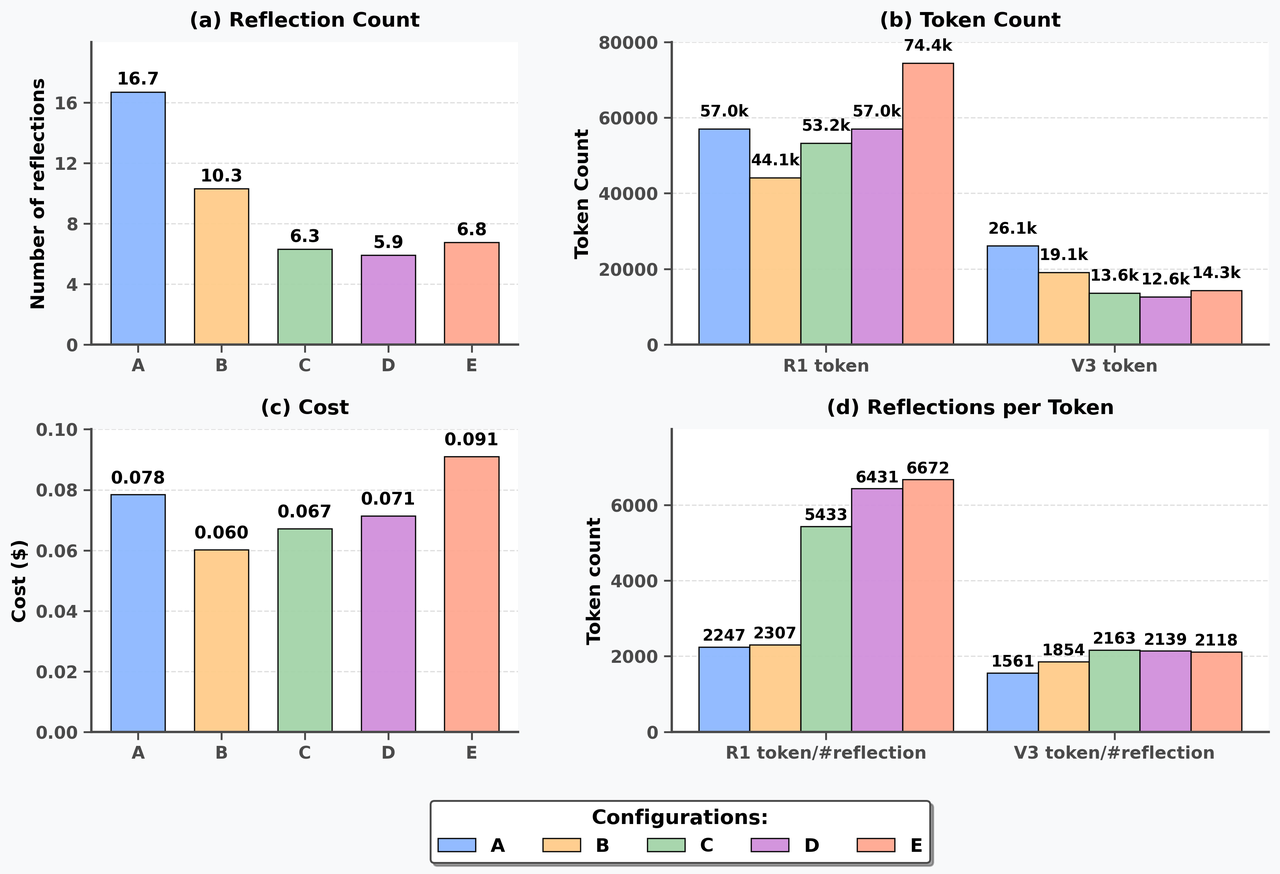}
\vspace{8 pt}
\caption{Results for NACA0012 Case 1 using Configurations A to E, showing averages of 10-step operational success runs. Metrics shown include: number of reflection iterations (\#reflection), total token consumption for \texttt{DeepSeek-R1} and \texttt{DeepSeek-V3} models (R1/V3 total token), execution cost (Cost \$), and average token consumption per reflection round for both models (R1/V3 token/\#reflection)}
\label{fig_naca0012_result1}
\end{figure}

Figure \ref{fig_naca0012_result1} presents a detailed analysis of the NACA0012 Case 1 performance across the five distinct Configurations (A through E), based on averaged results from 10-step operational success runs. This analysis yields several key insights into the system's operational efficiency and cost-effectiveness.

The evolution of reflection iterations, illustrated in Figure \ref{fig_naca0012_result1}(a), demonstrates an encouraging improvement pattern. As retrieve modules were integrated from configuration A to C, we observed a substantial decrease in required reflection iterations, followed by stabilization. This observation aligns with MetaOpenFOAM's findings \cite{chen2025metaopenfoam} regarding RAG positive impact on LLM response quality and CFD agent performance. The subsequent stability in reflection counts from configurations C to E suggests that additional RAG implementations, while beneficial for other aspects, had minimal impact on ChatCFD's reflection behavior. This plateau is reasonable given that \texttt{ContextRetriever} and comprehensive paper interpretation modules were primarily designed to enhance performance on physical-coupled cases and setup accuracy respectively, rather than affecting reflection behaviors for weak coupling cases.

An analysis of token consumption and associated costs, depicted in Figure \ref{fig_naca0012_result1}(b), reveals a pattern of improving resource utilization efficiency. Transitioning from Configuration A to D, while the token consumption of the \texttt{DeepSeek-R1} model remained relatively stable, the \texttt{DeepSeek-V3} model's consumption decreased markedly. This reduction contributed to lower per-case operational costs (decreasing from \$0.078 to \$0.071). This cost improvement, concurrent with a significant increase in operational success rates from 20\% to 80\%, underscores the enhanced overall effectiveness of the ChatCFD system. 

The transition from Configuration D to E resulted in a notable enhancement of system accuracy, as illustrated in Figure~\ref{fig_model_fusion_analyses}. This advancement was primarily achieved by improving the paper interpretation module's efficacy in extracting CFD case setups during Stage 2 and by ensuring greater consistency of these setups throughout the file correction processes in Stage 3. Although this upgrade led to increased \texttt{DeepSeek-R1} token consumption and a corresponding 30\% rise in operational cost (from \$0.071 to \$0.091), the substantial improvement in accurate case configuration—from 0\% to 40\%—justifies this investment. 

Figure \ref{fig_naca0012_result1}(c) illustrates the token consumption patterns per reflection iteration for both \texttt{DeepSeek-R1} and \texttt{DeepSeek-V3} models across the different Configurations. From Configuration A to E, the \texttt{DeepSeek-R1}'s consumption increases to more than two times, where the increasing consumption are mostly due to the integration of the \texttt{ContextRetriever} module in Configuration C. The \texttt{DeepSeek-V3}'s consumption initially rose due to expanded error handling capabilities and reference case integration through \texttt{ReferenceRetriever}. The subsequent stabilization of \texttt{DeepSeek-V3} token usage from Configuration C to E reflects the maturation of the reflection module's design, achieving optimal operational efficiency without necessitating additional \texttt{DeepSeek-V3} module invocations.
\begin{figure}[h!]
    \centering
    \includegraphics[width=6.25in]{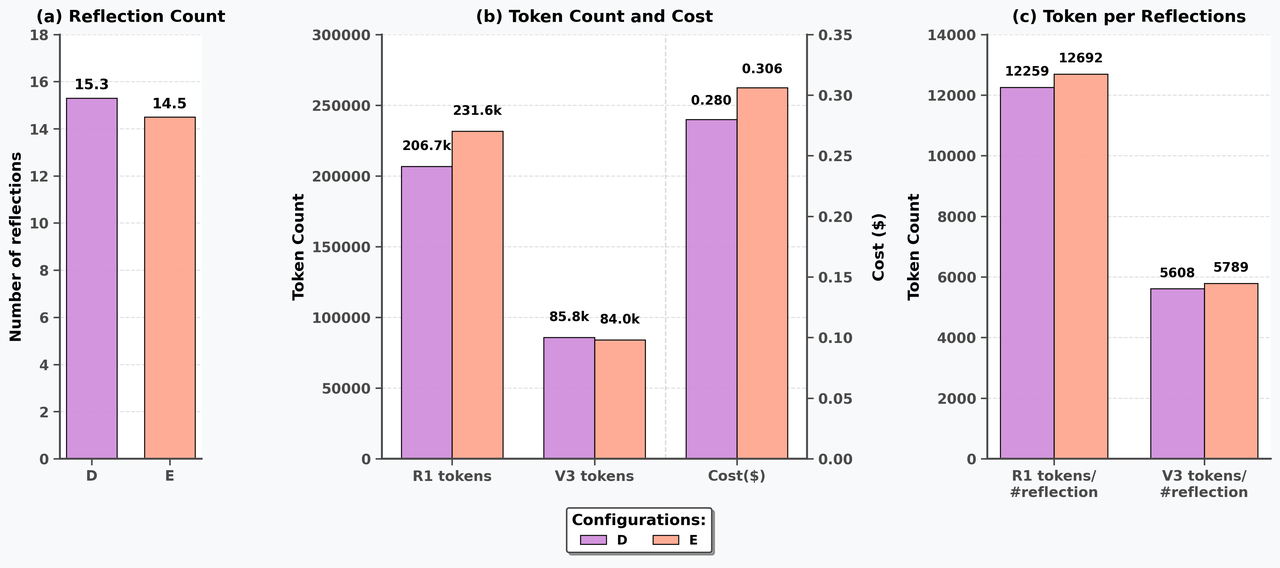}
    \caption{Results for Nozzle Case 5 using Configurations D and E, showing averages of 10-step operational success runs. Metrics shown include: number of reflection iterations (\#reflection), total token consumption for \texttt{DeepSeek-R1} and \texttt{DeepSeek-V3} models (R1/V3 total token), execution cost (Cost \$), and average token consumption per reflection round for both models (R1/V3 token/\#reflection)}
    \label{fig_nozzle_result1}
\end{figure}

Figure \ref{fig_nozzle_result1} presents a comparative performance analysis for the Nozzle Case 5, specifically contrasting Configurations D and E, based on simulations that successfully completed ten steps. Configurations A through C were omitted from this comparison due to their inability to consistently reach this ten-step benchmark. The key differentiator between Configurations D and E is the integration of the comprehensive paper interpretation module's enhanced article interpretation capabilities in Configuration E. Notwithstanding this enhancement, Figure \ref{fig_nozzle_result1}(a) indicates that the number of reflection iterations remained largely comparable between these two configurations, with only marginal reductions observed for Configuration E.

The deployment of comprehensive paper interpretation module resulted in increased token utilization, as depicted in Figure \ref{fig_nozzle_result1}(b). Configuration E demonstrated higher token consumption by the \texttt{DeepSeek-R1} model, whereas the \texttt{DeepSeek-V3} model's token usage remained stable across both Configurations D and E. In terms of operational cost, the average per-case expenditure saw a modest increase from \$0.28 to \$0.31, an approximate 10\% increment. Nevertheless, this additional investment yielded significant improvements in performance, particularly in case setup accuracy, which, as shown in Figure \ref{fig_model_fusion_analyses}, rose substantially from 0\% to 30\%.

Further examination, illustrated in Figure \ref{fig_nozzle_result1}(c), reveals that the average token consumption per reflection iteration for both the \texttt{DeepSeek-R1} and \texttt{DeepSeek-V3} models was comparable across Configurations D and E. This consistency underscores the efficacy of the system's architectural design, indicating that the comprehensive paper interpretation module detailed interpretation component operates as a distinct layer that augments overall system performance without adversely affecting the fundamental reflection mechanisms.

\begin{table}[htbp]
\caption{Statistical averages of accurate configured cases for Naca0012 Case 1 and Nozzle Case 5.}
\centering
\begin{tabular}{p{0.15\textwidth}p{0.35\textwidth}p{0.35\textwidth}}
\toprule
\textbf{Metrics} & \textbf{Naca0012, Case 1} & \textbf{Nozzle, Case 5} \\
\midrule
Physical models & 
\begin{tabular}{l}
    Flow type: Incompressible \\
    Solver: \texttt{simpleFoam} \\
    Turbulence: Spalart-Allmaras
\end{tabular}
& 
\begin{tabular}{l}
    Flow type: Compressible \\
    Solver: \texttt{rhoCentralFoam} \\
    Turbulence: Spalart-Allmaras \\
    Thermo model: hePsiThermo
\end{tabular} \\ \midrule
Number of configuration files & 9 & 12 \\ \midrule
Average tokens for all configuration files per case & 1,792 & 2,647 \\ \midrule
Configuration files & 
\parbox[t]{0.35\textwidth}{
\begin{tabular}{l}
    \texttt{0/p} \\
    \texttt{0/U} \\ 
    \texttt{0/nut} \\
    \texttt{0/nuTilda} \\
    \texttt{constant/transportProperties} \\
    \texttt{constant/turbulenceProperties} \\
    \texttt{system/controlDict} \\
    \texttt{system/fvSchemes} \\
    \texttt{system/fvSolution} \\
    \end{tabular}
}
& 
\parbox[t]{0.35\textwidth}{
\begin{tabular}{l}
    \texttt{0/p} \\
    \texttt{0/U} \\ 
    \texttt{0/nut} \\
    \texttt{0/nuTilda} \\
    \texttt{0/T} \\
    \texttt{0/alphat} \\
    \texttt{constant/transportProperties} \\
    \texttt{constant/turbulenceProperties} \\
    \texttt{constant/thermodynamicProperties} \\
    \texttt{system/controlDict} \\
    \texttt{system/fvSchemes} \\
    \texttt{system/fvSolution} \\
    \end{tabular}
}\\ \midrule
\texttt{DeepSeek-R1} call count & 19.1 & 48.3 \\ \midrule
\texttt{DeepSeek-V3} call count & 17.4 & 35.2 \\
\bottomrule
\end{tabular}
\label{tab:2}
\end{table}

\section{Appendix: CFD results of Cases 1 to 5}
\label{sec:A4}

\begin{figure}[h!]
\centering
\includegraphics[width=4in]{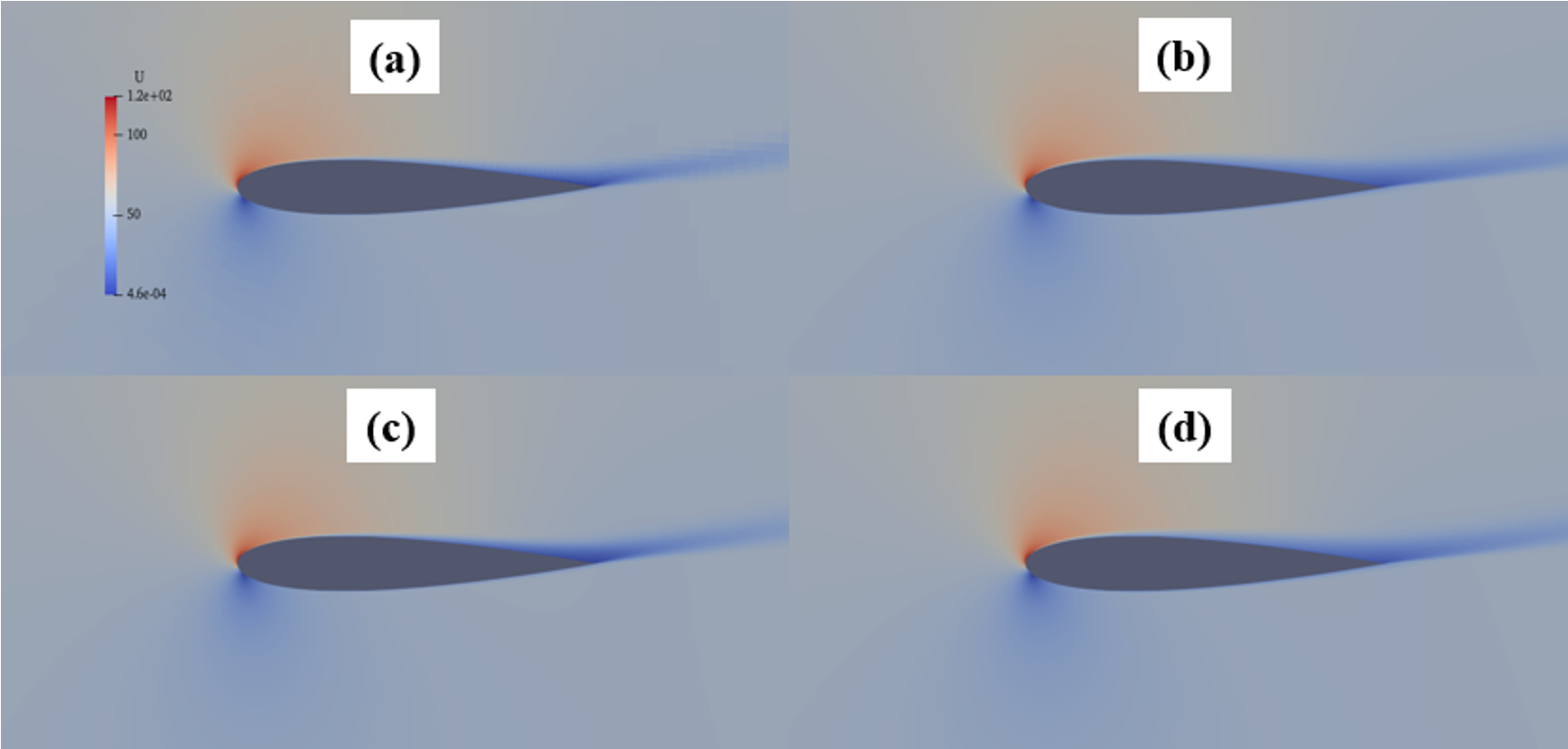}
\caption{Velocity magnitude contours for NACA0012 Cases 1 to 4 at 10° angle of attack using four turbulence models. (a) Spalart-Allmaras model, (b) k-$\epsilon$ model, (c) k-$\omega$ SST model, (d) RNGk-$\epsilon$ model.}
\label{naca_flowfield}
\end{figure}

\begin{figure}[h!]
\centering
\includegraphics[width=4in]{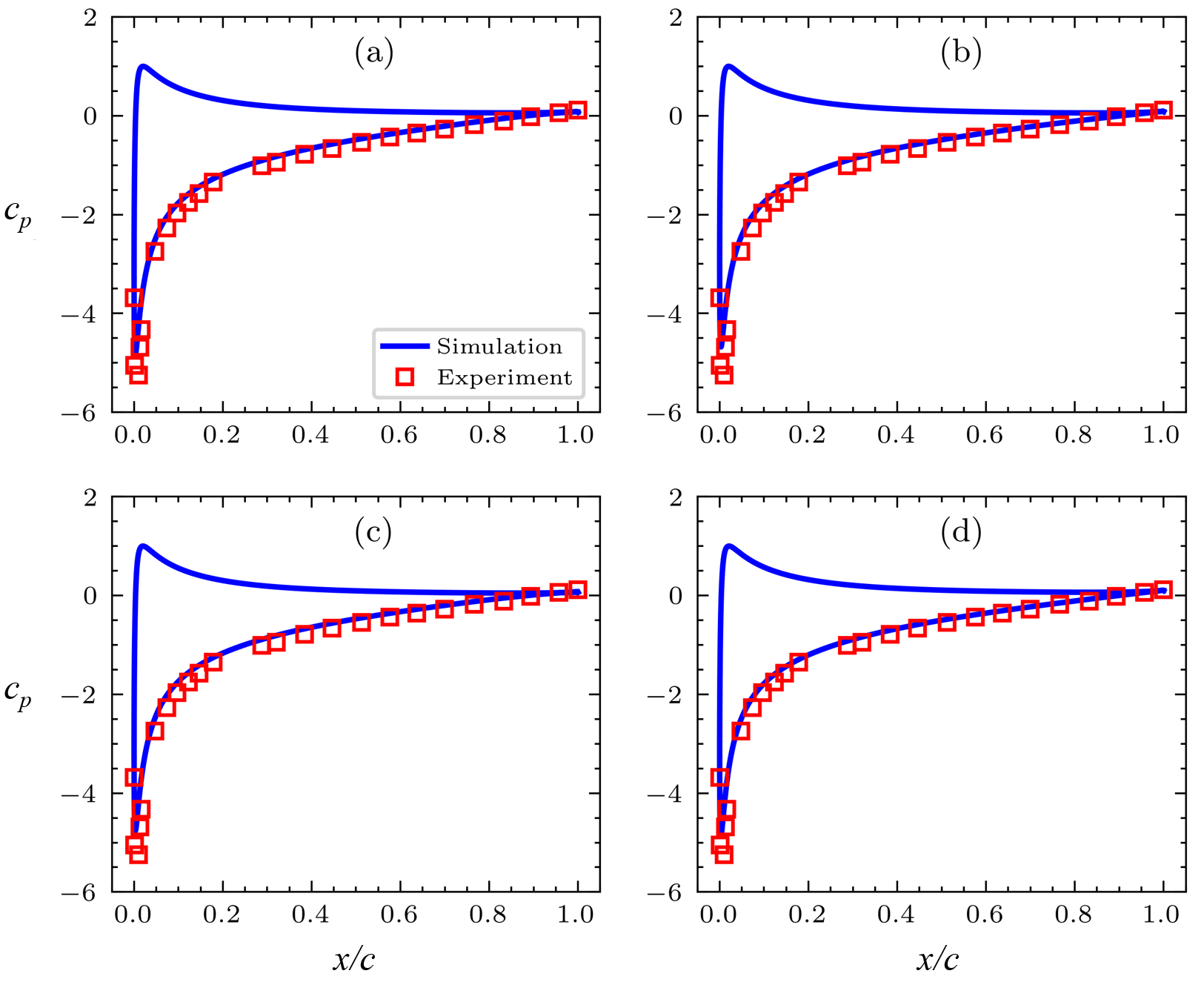}
\caption{Comparison of pressure coefficients between CFD simulation and experimental data for NACA0012 airfoil at 10° angle of attack, using (a) Spalart-Allmaras model, (b) k-$\epsilon$ model, (c) k-$\omega$ SST model, and (d) RNGk-$\epsilon$ model. Symbols represent experimental results reported by Gregory et al. \cite{gregory1970low}}
\label{naca_cp}
\end{figure}

Figures \ref{naca_flowfield} and \ref{naca_cp} show the velocity magnitude contours and surface pressure coefficient distributions for the NACA0012 airfoil Cases 1 to 4 at 10° angle of attack using different turbulence models. As can be seen, the velocity contours from the four cases are similar, and the pressure coefficient distributions agree well with the experimental results \cite{gregory1970low}.

\begin{figure}[h!]
\centering
\includegraphics[width=3in]{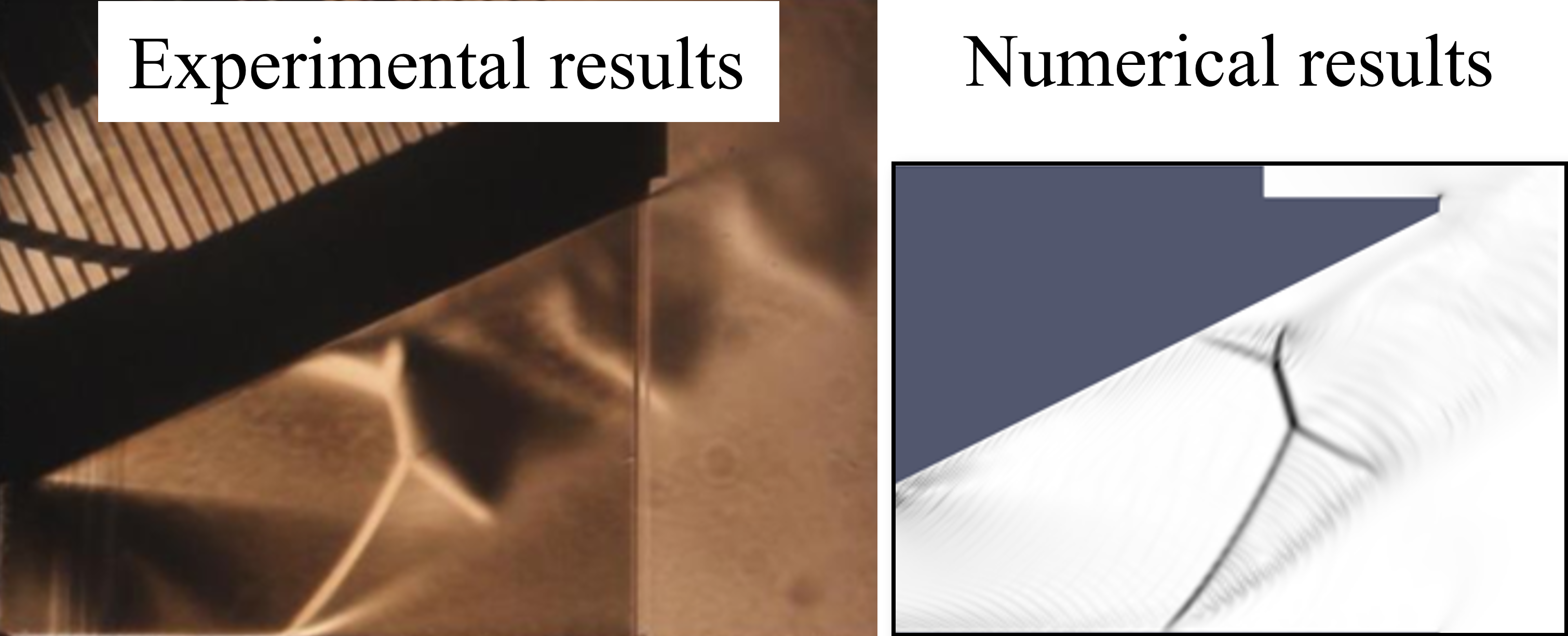}
\caption{(a) Experimental schlieren image \cite{yu2023comparative} and (b) numerical schlieren image of Case 5 for the Nozzle case at nozzle pressure ratio = 3.}
\label{nozzle_schlieren}
\end{figure}

Figure \ref{nozzle_schlieren} illustrates the experimental and numerical schlieren images for the nozzle case at a nozzle pressure ratio of 3. The experimental results are reported in Ref. \cite{yu2023comparative}. The numerical results exhibit a strong correspondence with the experimental observations, accurately capturing key flow features. The figure clearly depicts the formation of a Mach stem, resulting from the intersection and reflection of shock waves from the ramp and flap. This interaction also reveals the characteristic Mach reflection, distinguished by its $\lambda$-shock wave structure.

\end{document}